\documentclass[10pt,twocolumn,letterpaper]{article}
\PassOptionsToPackage{dvipsnames,table}{xcolor}

\usepackage{tikz}
\usetikzlibrary{
  calc, positioning, shapes, arrows.meta, backgrounds,
  decorations.pathreplacing, shapes.geometric
}

\usepackage[pagenumbers]{cvpr}

\definecolor{cvprblue}{rgb}{0.21,0.49,0.74}
\usepackage[pagebackref,breaklinks,colorlinks,allcolors=cvprblue]{hyperref}

\def\paperID{13850} 
\def\confName{CVPR}
\def\confYear{2025}

\title{PACT: Pruning and Clustering-Based Token Reduction for Faster Visual Language Models}

\author{
Mohamed Dhouib\\
LIX, École Polytechnique, IP Paris, France\\
{\tt\small mohamed.dhouib@polytechnique.edu}
\and
Davide Buscaldi\\
LIPN, Université Sorbonne Paris Nord, France\\
{\tt\small davide.buscaldi@lipn.univ-paris13.fr}
\and
Sonia Vanier\\
LIX, École Polytechnique, IP Paris, France\\
{\tt\small sonia.vanier@polytechnique.edu}
\and
Aymen Shabou\\
DataLab Groupe, Crédit Agricole S.A, France\\
{\tt\small aymen.shabou@credit-agricole-sa.fr}
}

\usepackage{algorithm, algpseudocode, amsmath}

\usepackage[pagebackref,breaklinks,colorlinks]{hyperref}
\algnewcommand\INPUT{\item[\textbf{Input:}]}
\algnewcommand\OUTPUT{\item[\textbf{Output:}]}
\usepackage{amsmath} 
\usepackage{amssymb}
\usepackage{threeparttable}

\begin{document}
\maketitle
\begin{abstract}
Visual Language Models require substantial computational resources for inference due to the additional input tokens needed to represent visual information. However, these visual tokens often contain redundant and unimportant information, resulting in an unnecessarily high number of tokens. To address this, we introduce \textbf{PACT}, a method that reduces inference time and memory usage by pruning irrelevant tokens and merging visually redundant ones at an early layer of the language model. Our approach uses a novel importance metric to identify unimportant tokens without relying on attention scores, making it compatible with FlashAttention. We also propose a novel clustering algorithm, called Distance Bounded Density Peak Clustering, which efficiently clusters visual tokens while constraining the distances between elements within a cluster by a predefined threshold. We demonstrate the effectiveness of \textbf{PACT} through extensive experiments.
\end{abstract}    

\section{Introduction}

Extending Large language models to modalities other than text~\cite{Video-llama,MMLLMS,onellm,Qwen-audio,kosmos} has seen success in recent years across various domains, especially in the visual domain with models like LLaVA~\cite{LLAVA} and Qwen-VL~\cite{QWEN-Vl}. State-of-the-art Visual Language Models generally consist of three main components: a vision encoder, a connector, and a language model. The vision encoder converts input images into visual tokens, which are passed through the connector and then fed to the language model along with the input text. While this architecture has shown impressive performance across different tasks, it suffers from high computational cost due to the large number of visual tokens. In this paper, we introduce two complementary methods to optimize Visual Language Models by reducing inference time and memory requirements: a pruning module and a clustering algorithm. These methods can be used independently or combined, forming the \textbf{PACT} approach for greater effectiveness. Notably, our pruning and clustering modules, as well as \textbf{PACT}, are applied at inference time and thus require no additional training. The pruning module identifies unimportant visual tokens based on a novel importance metric that evaluates each token's relevance without relying on attention scores. This makes it compatible with FlashAttention~\cite{flashattention}, as FlashAttention does not support the calculation of attention scores. The second module introduces a novel clustering algorithm, \textbf{Distance Bounded Density Peak Clustering (\textbf{DBDPC})}, which clusters visual tokens while ensuring that the distances between elements within a cluster are constrained by a predefined threshold. By combining these two methods, we develop \textbf{PACT}. First, the pruning module eliminates unimportant tokens, then the \textbf{DBDPC} algorithm clusters the remaining ones. Tokens that were initially pruned but are sufficiently close to the constructed clusters are reincorporated, ensuring that valuable information from the pruned tokens is recovered. Finally, the tokens within each cluster are merged into a single representative token, reducing the total token count.\\ By combining both pruning and clustering, \textbf{PACT} achieves an effective visual token reduction, addressing both irrelevant and redundant tokens. When applied to LLaVA-OneVision-7B, \textbf{PACT} achieves a 50\% visual token reduction with negligible performance loss. Moreover, \textbf{PACT} exhibits significantly less performance degradation at higher reduction ratios compared to previous methods, achieving 71.3\% visual token reduction ratio with only 1.4\% performance drop, whereas previous state-of-the-art methods show at best a 4.4\% performance drop at an equal reduction ratio.
Our contributions are as follows:
\begin{itemize}
    \item We propose a novel visual token pruning metric that does not rely on attention scores, ensuring compatibility with FlashAttention, and empirically validate its effectiveness.
    \item We introduce a new clustering algorithm aimed at reducing visual redundancy and show its superiority over other clustering algorithms for visual token reduction.
     \item We show that combining pruning with clustering-based merging surpasses either technique alone for visual token reduction. By integrating our pruning and clustering algorithms, we propose a novel approach, \textbf{PACT}, and demonstrate that it outperforms previous and concurrent works \cite{fastv,tome,tokenwith,hired,llavapurge}. The codebase used to obtain the results in this study is available at \href{https://github.com/orailix/PACT/tree/main}{\textcolor{blue}{https://github.com/orailix/PACT/tree/main}}.
\end{itemize}

\section{Related work} 
\subsection{Visual language models}
Since the introduction of BLIP-2 \cite{Blip-2}, the use of a visual encoder followed by a connector that feeds visual vectors to the language model has become the standard architecture for Visual Language Models (VLMs) \cite{OWL,minigpt4,Shikra}.
Recent models \cite{llavaone,internvl,Qwen2-VL} have enhanced VLM architecture with high-resolution handling, which is necessary for document understanding tasks \cite{donut,docparser}. LLaVA-OneVision \cite{llavaone} divides images into 384$\times$384 crops, encodes each part with SigLIP \cite{siglip}, and uses bilinear interpolation to reduce token count up to 8,748 tokens. InternVL2 \cite{internvl} splits images into 448$\times$448 tiles, processing up to 40 tiles per image with InternViT~\cite{internvl}, and applies pixel shuffle to reduce the number of visual tokens, producing up to 10,240 tokens. Qwen-VL2 \cite{Qwen2-VL} uses 2D Rotary Positional Embeddings for dynamic resolution support and merges adjacent tokens via an MLP layer, yet still requires over 10,000 tokens for high resolution images. While these models apply token reduction by merging adjacent tokens to preserve structure, they do not address token irrelevance or redundancy, limiting efficiency.

\subsection{Visual token reduction}
Reducing the number of visual tokens in Vision Transformers (ViT) has been a key focus of the research community for several years. EViT \cite{EViT} identifies and merges irrelevant tokens by relying on the attention scores between the class token (\verb|[CLS]|) and visual tokens. ToME \cite{tome} proposed a simple yet effective approach that iteratively merges similar tokens throughout the ViT layers. Building on these ideas, recent efforts have extended visual token reduction techniques to VLMs. LaVIT \cite{lavit} used the Gumbel-Softmax \cite{gumbelsoftmax} to train a mask that selects tokens for retention, merging discarded tokens into retained ones via additional attention layers. LLaVA-PruMerge \cite{llavapurge} accelerates LLAVA 1.5 \cite{LLAVA} by leveraging the attention scores between the \verb|[CLS]| token and visual tokens in the last layer of the ViT encoder to decide which tokens to retain, while HiRED \cite{hired} refines this approach by allocating token budgets based on attention from earlier layers. However, both these methods are only applicable to architectures where a ViT is used and a \verb|[CLS]| token is added to the input visual sequence, making them incompatible with the majority state-of-the-art VLMs, which do not use a \verb|[CLS]| token. Moreover, both methods attribute scores to tokens at the output of the visual encoder, but recent VLMs merge adjacent visual tokens before passing them to the language model. It is unclear how to attribute pre-merging scores to the resulting tokens, making LLaVA-PruMerge and HiRED inapplicable. We note that LLaVA-PruMerge mitigates information loss by merging pruned tokens with retained ones. However, it does not merge similar retained tokens; therefore, it does not address visual redundancy, a typical limitation of pruning-based methods. TRIM \cite{lessismoretext} prunes tokens based on similarity with pooled text from CLIP \cite{clip}. However, as TRIM relies on textual information for pruning, it is less suitable for multi-turn conversations where, in practice, visual tokens would be pruned solely based on the text information available during the image’s forward pass, potentially losing crucial information required to answer subsequent prompts. FastV \cite{fastv} evaluates token importance via average attention scores, which is not compatible with FlashAttention, adding computational overhead for recent VLMs. VTW \cite{tokenwith} removes tokens in deeper layers. While this method shows promising results, its reduction of computational costs is limited as visual tokens are only withdrawn in later layers. \\ These previous methods address only one of two issues: the presence of unimportant tokens or visual redundancy. In this work, we introduce \textbf{PACT}, a novel approach that tackles both issues simultaneously by pruning irrelevant tokens and merging visually redundant ones.
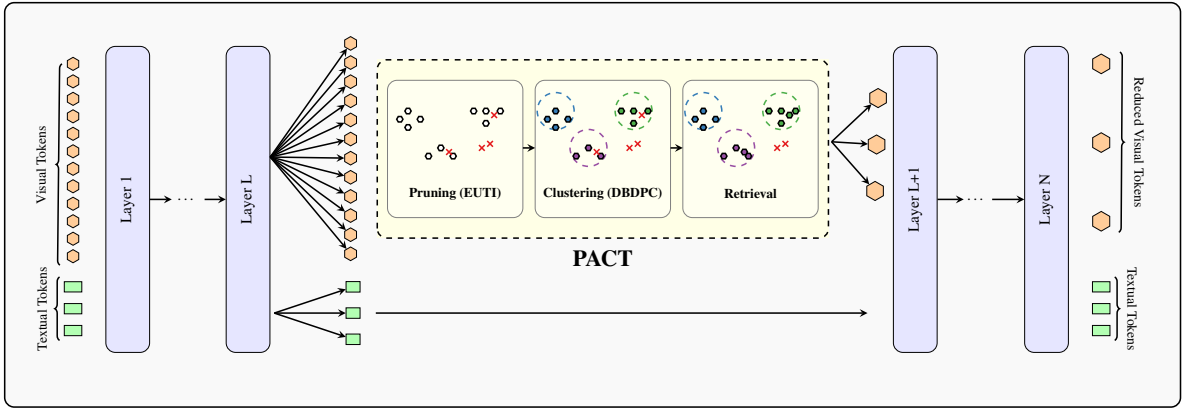
\begin{figure*}[h!]
    \centering
    \resizebox{0.9\textwidth}{!}{\begin{tikzpicture}[>=stealth]

\begin{scope}[on background layer]
   \draw[fill=gray!05!white, draw=black, line width=0.8pt, rounded corners=5pt]
    (-5.25, -4.6) rectangle (16.25, 2.8);
\end{scope}

\begin{scope}[scale=0.8, every node/.style={transform shape}]
  \foreach \i in {0,...,11} {
    \node[
      draw,
      regular polygon,
      regular polygon sides=6,
      fill=orange!40,
      rotate=90,
      minimum size=0.3cm,
      inner sep=0pt,
      line width=0.5pt
    ] (visL\i) at (-5, {2.1 - 0.4*\i}) {};
  }
  \draw[decorate,decoration={brace,mirror},thick]
    ($(visL0.north west)+(-0.15,0.1)$)
    -- ($(visL11.south west)+(-0.4,-0.1)$)
    node[midway,left=0.45cm,rotate=90,anchor=center]{\small Visual Tokens};

  \node[draw,rectangle,fill=green!30,minimum width=0.4cm,minimum height=0.2cm]
    (txtL1) at (-5,-3.0) {};
  \node[draw,rectangle,fill=green!30,minimum width=0.4cm,minimum height=0.2cm]
    (txtL2) at (-5,-3.5) {};
  \node[draw,rectangle,fill=green!30,minimum width=0.4cm,minimum height=0.2cm]
    (txtL3) at (-5,-4.) {};
  \draw[decorate,decoration={brace,mirror},thick]
    ($(txtL1.north west)+(-0.1,0.1)$)
    -- ($(txtL3.south west)+(-0.1,-0.1)$)
    node[midway,left=0.4cm,rotate=90,anchor=center]{\small Textual Tokens};

  \begin{scope}[scale=0.8, every node/.style={transform shape},xshift=9.35cm,yshift=-1cm]
    \node[draw,rectangle,fill=green!30,minimum width=0.4cm,minimum height=0.3cm]
    (txtL1) at (-7.6,-2.75) {};
  \node[draw,rectangle,fill=green!30,minimum width=0.4cm,minimum height=0.3cm]
    (txtL2) at (-7.6,-3.5) {};
  \node[draw,rectangle,fill=green!30,minimum width=0.4cm,minimum height=0.3cm]
    (txtL3) at (-7.6,-4.25) {};
 \coordinate (o) at (-9.85,-3.5);
\draw[->,thick] (o) --  (txtL1) ;
\draw[->,thick] (o) --  (txtL2) ;
\draw[->,thick] (o) --  (txtL3) ;
    \draw[->,thick] (-6.95,-3.5) --  (7.1,-3.5) ;
    \end{scope}

  \usetikzlibrary{shadows}

  \node[
  draw, fill=blue!10,
  minimum width=1.0cm,
  minimum height=7cm,
  rounded corners=5pt,
  font=\normalsize
  ] (L1) at (-3.75,-1) {\rotatebox{90}{Layer 1}};

  \node at (-2.4,-1) {\dots};

   \node[
  draw, fill=blue!10,
  minimum width=1.0cm,
  minimum height=7cm,
  rounded corners=5pt,
  font=\normalsize
  ] (LL) at (-1,-1) {\rotatebox{90}{Layer L}};

   \node[
  draw, fill=blue!10,
  minimum width=1.0cm,
  minimum height=7cm,
  rounded corners=5pt,
  font=\normalsize
  ] (LLp) at (14.25,-1) {\rotatebox{90}{Layer L+1}};

  \node at (15.65,-1) {\dots};

   \node[
  draw, fill=blue!10,
  minimum width=1.0cm,
  minimum height=7cm,
  rounded corners=5pt,
  font=\normalsize
  ] (LN) at (17.25,-1) {\rotatebox{90}{Layer N}};

  \draw[->,thick] (L1.east) -- ++(0.5,0);
  \draw[->,thick] ($(L1.east) + (1.25,.0)$) -- (LL.west);

  \draw[->,thick] (LLp.east) -- ++(0.6,0);
  \draw[->,thick] ($(LLp.east) + (1.25,.0)$) -- (LN.west);

  \foreach \i in {0,...,2} {
    \node[
      draw,
      regular polygon,
      regular polygon sides=6,
      fill=orange!40,
      rotate=90,
      minimum size=0.45cm,
      inner sep=0pt,
      line width=0.5pt
    ] (visR\i) at (18.5, {2.1 - 1.8*\i}) {};
  }
  \draw[decorate,decoration={brace},thick]
    ($(visR0.north east)+(0.575,0.12)$)
    -- ($(visR2.south east)+(0.25,-0.4)$)
    node[midway,right=0.45cm,rotate=270,anchor=center]{\small Reduced Visual Tokens};

  \node[draw,rectangle,fill=green!30,minimum width=0.4cm,minimum height=0.2cm]
    (txtR1) at (18.5,-3.0) {};
  \node[draw,rectangle,fill=green!30,minimum width=0.4cm,minimum height=0.2cm]
    (txtR2) at (18.5,-3.5) {};
  \node[draw,rectangle,fill=green!30,minimum width=0.4cm,minimum height=0.2cm]
    (txtR3) at (18.5,-4.0) {};
  \draw[decorate,decoration={brace},thick]
    ($(txtR1.north east)+(0.1,0.1)$)
    -- ($(txtR3.south east)+(0.1,-0.1)$)
    node[midway,right=0.4cm,rotate=270,anchor=center]{\small Textual Tokens};
\end{scope}

\path let \p1 = (LL.east) in coordinate (pactWest) at (0.25,{\y1 + 30.0});


\foreach \i in {0,...,11} {
  \node[draw,
      regular polygon,
      regular polygon sides=6,
      fill=orange!40,
      rotate=90,
      minimum size=0.25cm,
      inner sep=0pt,
      line width=0.3pt]
        (pactIn\i) at ($(pactWest)+(0.825, 1.8 - 0.35*\i)$) {};
  \draw[->,thick] ($(LL.east)+(0,0.775)$) -- (pactIn\i);
}

\begin{scope}[shift={(pactWest)}, scale=0.75]
  \definecolor{prunedcolor}{RGB}{228,26,28}
  \definecolor{cluster1}{RGB}{55,126,184}
  \definecolor{cluster2}{RGB}{77,175,74}
  \definecolor{cluster3}{RGB}{152,78,163}
  \definecolor{softyellow}{RGB}{255,255,240}
  \definecolor{darkeryellow}{RGB}{255,255,230}

  \tikzstyle{token} = [
    regular polygon,
    regular polygon sides=6,
    draw=black,
    thick,
    inner sep=1pt,
    fill=white
  ]
  \tikzstyle{tokenA}      = [token, fill=cluster1]
  \tikzstyle{tokenB}      = [token, fill=cluster2]
  \tikzstyle{tokenC}      = [token, fill=cluster3]
  \tikzstyle{tokenPruned} = [token, draw=prunedcolor, fill=white]

  \tikzstyle{stepbackground} = [
    draw=gray,
    fill=softyellow,
    rounded corners=5pt
  ]
  \tikzstyle{bottomtitle} = [
    font=\bfseries\scriptsize,
    text=black
  ]

  \newcommand{\prunedCross}[1]{%
    \draw[prunedcolor, thick]
      ($(#1)+(-0.07,-0.07)$) -- ($(#1)+(0.07,0.07)$);
    \draw[prunedcolor, thick]
      ($(#1)+(-0.07,0.07)$) -- ($(#1)+(0.07,-0.07)$);
  }

  \begin{scope}[on background layer,yshift=-2.25cm]
    \fill[darkeryellow]
      (2 - 0.25, 0.15 - 0.25)
      rectangle
      (12.5 + 0.25, 4 + 0.25);
    \draw[black, dashed, thick, rounded corners=5pt]
      (2 - 0.25, 0.15 - 0.25)
      rectangle
      (12.5 + 0.25, 4 + 0.25);
  \end{scope}

  \begin{scope}[shift={(2,0)},yshift=-2.25cm]
    \begin{scope}[on background layer]
      \path[stepbackground] (0,0.4) rectangle (3.3,3.75);
    \end{scope}
    \node[bottomtitle] at (1.65,0.98) {Pruning (EUTI)};

    \foreach \i/\x/\y in {
      1/0.3/2.8,  2/0.5/3.0,  3/0.8/2.8,  4/0.5/2.6,
      5/2.1/3.0,  6/2.4/3.0,  7/2.75/3.0, 8/2.4/2.7,
      9/1.0/1.9, 10/1.3/2.1,  11/1.6/1.9,
      12/2.6/2.9,13/1.5/2.0,
      14/2.5/2.2,15/2.3/2.1
    }{
      \coordinate (S1P\i) at (\x,\y);
    }

    \foreach \i in {1,...,11} {
      \node[token] at (S1P\i) {};
    }
    \foreach \i in {12,13,14,15} {
      \prunedCross{S1P\i}
    }
  \end{scope}

  \begin{scope}[shift={(5.6,0)},yshift=-2.25cm]
    \begin{scope}[on background layer]
      \path[stepbackground] (0,0.4) rectangle (3.3,3.75);
    \end{scope}
    \node[bottomtitle] at (1.65,0.98) {Clustering (DBDPC)};

    \foreach \i in {1,...,15} {
      \coordinate (S2P\i) at ($(S1P\i)+(3.6,0)$);
    }

    \draw[cluster1, dashed, thick] (S2P2) circle(0.45);
    \foreach \j in {1,2,3,4} {
      \node[tokenA] at (S2P\j) {};
    }
    \draw[cluster2, dashed, thick] (S2P6) circle(0.45);
    \foreach \j in {5,6,7,8} {
      \node[tokenB] at (S2P\j) {};
    }
    \draw[cluster3, dashed, thick] (S2P10) circle(0.45);
    \foreach \j in {9,10,11} {
      \node[tokenC] at (S2P\j) {};
    }
    \foreach \j in {12,13,14,15} {
      \prunedCross{S2P\j}
    }
  \end{scope}

  \begin{scope}[shift={(9.2,0)},yshift=-2.25cm]
    \begin{scope}[on background layer]
      \path[stepbackground] (0,0.4) rectangle (3.3,3.75);
    \end{scope}
    \node[bottomtitle] at (1.65,0.98) {Retrieval};

    \foreach \i in {1,...,15} {
      \coordinate (S3P\i) at ($(S1P\i)+(7.2,0)$);
    }

    \draw[cluster1, dashed, thick] (S3P2) circle(0.45);
    \foreach \j in {1,2,3,4} {
      \node[tokenA] at (S3P\j) {};
    }
    \draw[cluster2, dashed, thick] (S3P6) circle(0.45);
    \foreach \j in {5,6,7,8,12} {
      \node[tokenB] at (S3P\j) {};
    }
    \draw[cluster3, dashed, thick] (S3P10) circle(0.45);
    \foreach \j in {9,10,11,13} {
      \node[tokenC] at (S3P\j) {};
    }
    \foreach \j in {14,15} {
      \prunedCross{S3P\j}
    }
  \end{scope}

  \draw[->, thick] (5.3,-0.15) -- (5.6,-0.15);
  \draw[->, thick] (8.9,-0.15) -- (9.2,-0.15);

  \node[font=\bfseries\large, text=black] at (7.25,-2.85) {PACT};
\end{scope}

\path let \p1 = (pactWest) in coordinate (pactEast) at (9.87,\y1);

\foreach \i in {0,...,2} {
  \node[draw,
      regular polygon,
      regular polygon sides=6,
      fill=orange!40,
      rotate=90,
      minimum size=0.35cm,
      inner sep=0pt,
      line width=0.3pt]
        (pactOut\i) at ($(pactEast)+(0.85-0.0275*\i, 0.8 - 0.85*\i)$) {};
  \draw[->,thick] (pactEast) -- (pactOut\i);
}

\end{tikzpicture}}
    \caption{\textbf{Simplified illustration of PACT.} This figure illustrates the three-step process of \textbf{PACT}:  
(1) First, \textbf{EUTI} is used to prune visual tokens deemed unimportant;  
(2) Then, \textbf{DBDPC} is applied to cluster the remaining tokens, ensuring that the distance between each token and its corresponding cluster center is smaller than the cutoff distance; 
(3) Finally, initially pruned tokens that are close to cluster centers are reintegrated, and the elements within each cluster are merged to form the reduced set of visual tokens.
}
    \label{fig:pact}
\end{figure*}
\section{Method}
\label{method}
In this section, we present \textbf{PACT}, a method that aims to reduce VLMs inference time and memory usage by pruning unimportant tokens and merging visually redundant ones at an early layer \( L \) of the language model. \textbf{PACT} consists of three steps: First, unimportant tokens are identified. Next, the remaining tokens are clustered. Finally, tokens in each cluster, along with sufficiently close tokens that were initially discarded, are merged. \textbf{PACT} operates within a selected layer \(L\) of the language model and is applicable in scenarios where visual tokens are fed into the language model, regardless of the architecture of the visual encoder or connector. The three-step process of PACT is illustrated in \Cref{fig:pact}. We denote the hidden states at layer \(L\) by \(\mathbf{H} \in \mathbb{R}^{n \times d}\), where \(n\) is the number of visual tokens and \(d\) is the dimensionality of the hidden states. We denote by \(\mathbf{K}, \mathbf{Q} \in \mathbb{R}^{n \times n_h \times d_h}\) the key and query matrices for the visual tokens at layer \(L\), where \(n_h\) represents the number of attention heads and \(d_h\) is the dimensionality of each attention head. For simplicity, we omit the layer index in the notation. We denote the position index of a token by a subscript, while the attention head is indicated by a superscript. For instance, \(\mathbf{k}_i^{(j)}\) represents the key vector corresponding to the \(i\)-th visual token and the \(j\)-th attention head. 
 
\subsection{Unimportant tokens identification}
\begin{algorithm}[!t]
\centering
\caption{ EUTI}
\label{algo:pruning}
\begin{algorithmic}[0]
\INPUT Hidden states $\mathbf{H} \in \mathbb{R}^{n \times d}$; key and query matrices $\mathbf{K}, \mathbf{Q} \in \mathbb{R}^{n \times n_h \times d_h}$; pruning percentage $\lambda \in [0, 1]$
\OUTPUT Sets of important and unimportant visual tokens

\State \textbf{Step 1: Calculate the global query vector}
\State $\mathbf{Q}_{\text{global}} = \frac{1}{n} \sum_{i=1}^{n} \mathbf{Q}_i$

\State \textbf{Step 2: Compute the importance score for each visual token}
\ForAll{$i = 1, \dots, n$}
    \State $s_i = \frac{1}{n_h} \sum_{j=1}^{n_h} Softmax(\mathbf{k}_i^{(j)} \cdot \mathbf{Q}_{\text{global}}^{(j)}) \cdot \|\mathbf{h}_i\|_2$
\EndFor

\State \textbf{Step 3: Define sets of important and unimportant tokens}
\State $S_{\text{important}} = \{ i \mid s_i \geq \text{Percentile}(s, \lambda) \}$
\State $S_{\text{unimportant}} = \{ i \mid s_i < \text{Percentile}(s, \lambda) \}$

\State \textbf{Return} $S_{\text{important}},\ S_{\text{unimportant}}$
\end{algorithmic}
\end{algorithm}
\begin{figure}[!b]
    \centering

    \begin{minipage}[t]{0.235\textwidth}
        \centering
        \includegraphics[width=\textwidth]{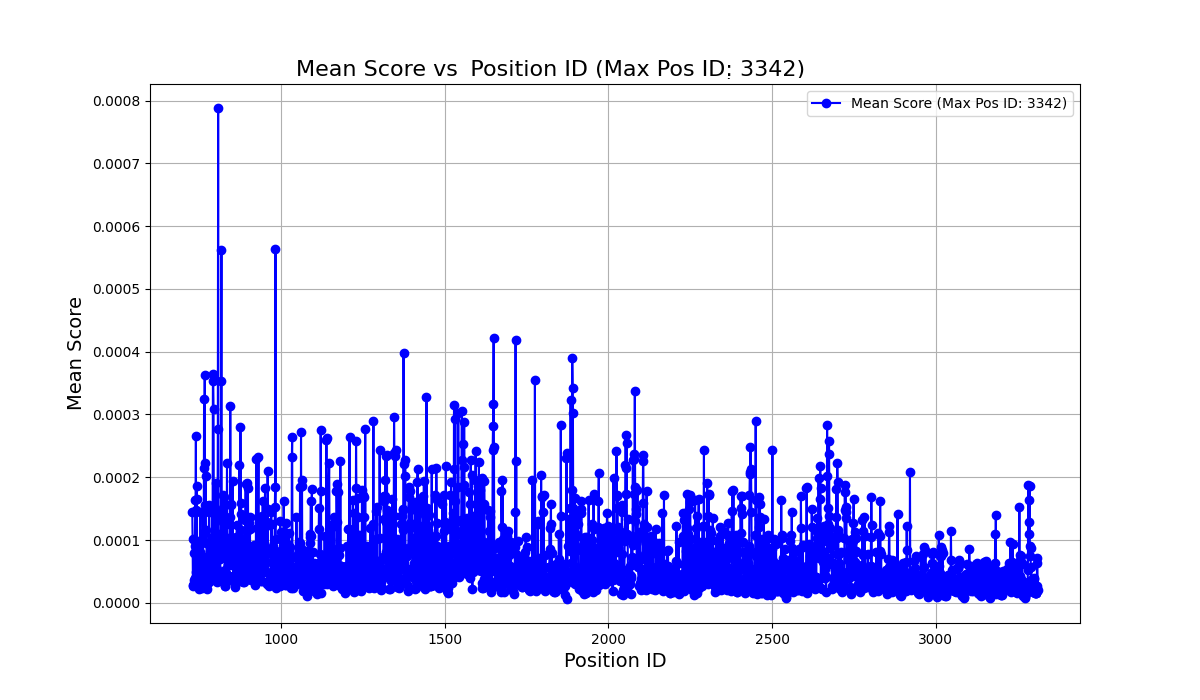}
        \caption*{(a) Average attention scores as a function of Position IDs.}
        \label{fig:mean_score_raw}
    \end{minipage}
    \hfill
    \begin{minipage}[t]{0.235\textwidth}
        \centering
        \includegraphics[width=\textwidth]{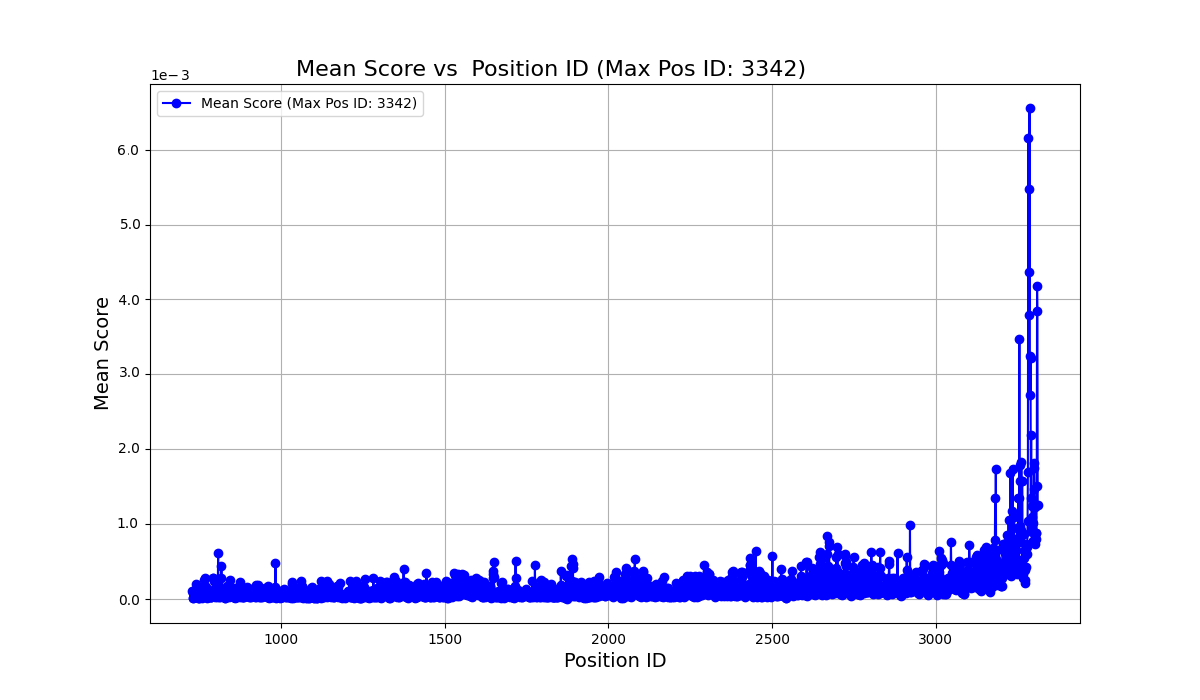}
        \caption*{(b) Average attention scores relative to non-masked tokens as a function of Position IDs.}
        \label{fig:mean_score_calibrated}
    \end{minipage}
    \caption{\textbf{Illustration of the bias induced by the use of the average attention scores across visual tokens as a pruning metric.} In (a), averaging attention over all tokens favors earlier tokens, leading to pruning later tokens more frequently. In (b), averaging only over attending tokens reverses the bias, leading to earlier tokens being pruned more often. 
    }
    \label{fig:attention_bias_comparison}
\end{figure} 
 A straightforward approach to identifying unimportant tokens at a certain layer $L$ of the used language model is to define the importance of each token as the total attention score that a given token receives from all other tokens \cite{fastv}. However, this method has three main drawbacks. First, current VLMs utilize FlashAttention \cite{flashattention}, which does not support outputting attention scores. Secondly, attention scores are computed with masking, which introduces biases. Tokens at the end of a sequence tend to receive lower average attention scores since fewer tokens attend to them. Calculating the average attention score for each token based solely on the tokens that attend to it can mitigate this masking effect but introduces a new bias: end-of-sequence tokens may exhibit higher scores as they receive attention mainly from nearby tokens. This leads to either earlier or later tokens being pruned more frequently, as shown in \cref{fig:attention_bias_comparison}. Such positional bias should be avoided, as pruning should depend solely on the information that visual tokens hold, not their position. Finally, relying only on keys and queries at a single layer to determine an importance metric may fail to fully capture the significance of visual tokens across all layers of the language model, mainly because each self-attention layer focuses on different aspects of the visual tokens. To address this, we propose an importance metric that incorporates the accumulated information from the hidden states and the layer-specific information from the keys and queries at an early layer \( L \). We refer to this method as \textbf{E}fficient \textbf{U}nimportant \textbf{T}okens \textbf{I}dentification (\textbf{EUTI}). We speculate that the norm of hidden states can provide critical information about the importance of each visual token, as they reflect how much information a particular token carries through the network. 
 \begin{figure}[!t]
    \centering
    \includegraphics[width=0.45\textwidth]{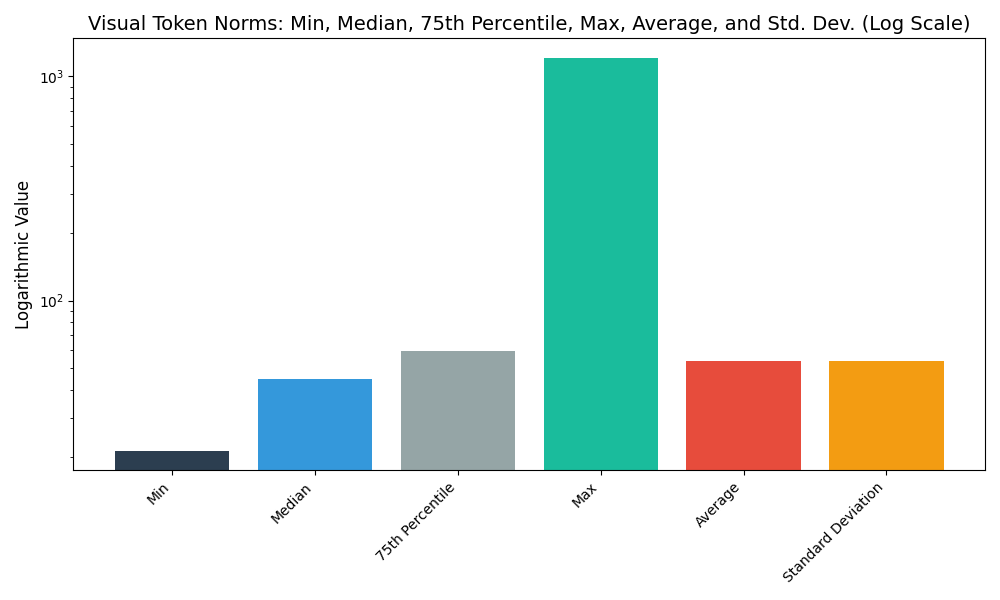}
    \caption{\textbf{Illustration of visual token norm statistics at the fourth layer of LLaVA-OneVision-7B.}}
    \label{fig:norm_analysis}
\end{figure} 
\begin{figure}[!b]
    \centering
    \includegraphics[width=0.45\textwidth]{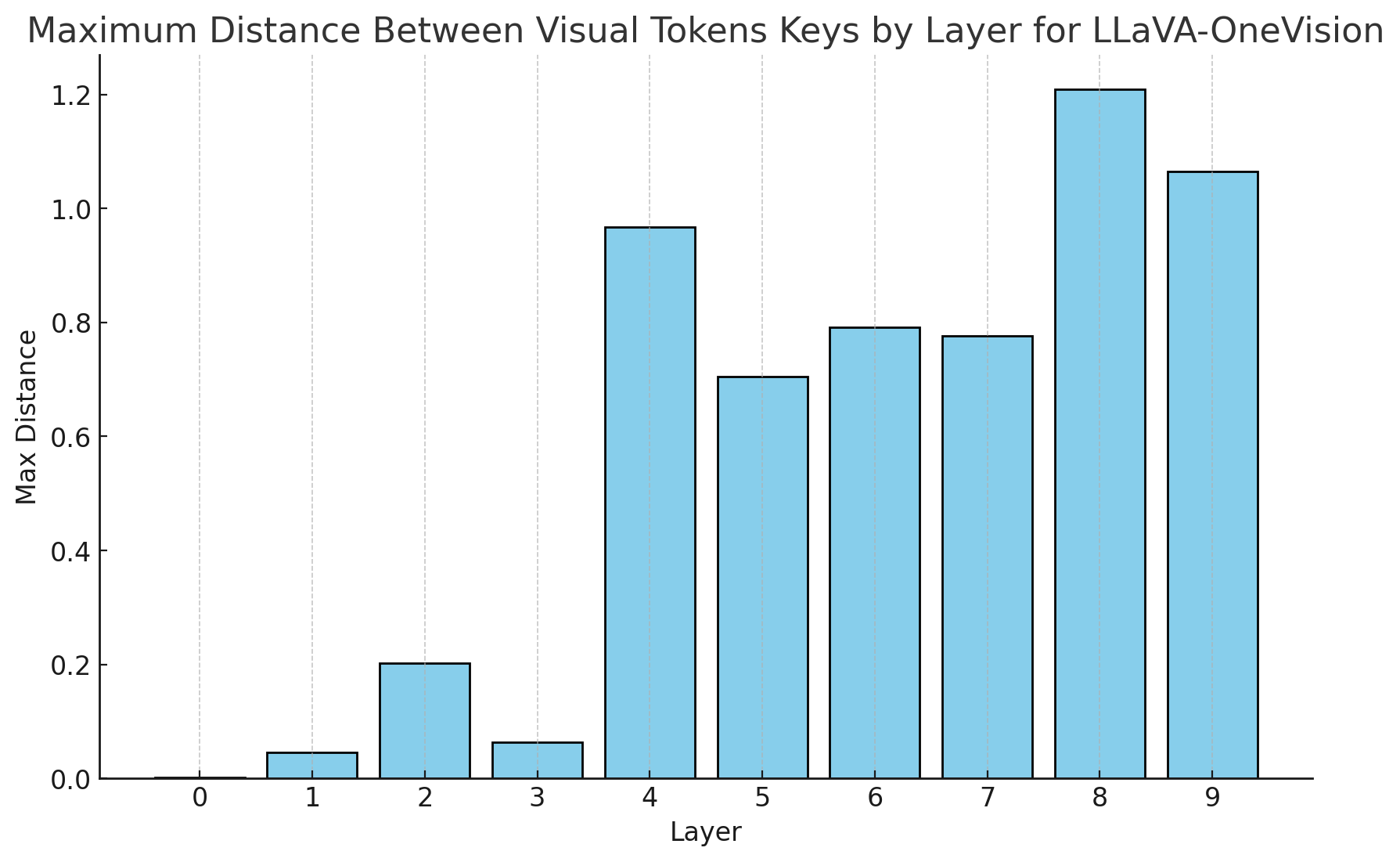}
\caption{\textbf{Illustration of the maximum distance between the keys of visual tokens for the first 10 layers of LLaVA-OneVision-7B before the application of rotary embeddings.}}
    \label{plotdistancekeys}
\end{figure} \Cref{fig:norm_analysis} presents statistics on the hidden state norms of visual tokens at the fourth layer of LLaVA-OneVision-7B, indicating a high variance. This variance suggests that certain visual tokens accumulate more information through residual connections and may therefore be more important for subsequent calculations. To leverage information from both hidden state norms  and the key and query vectors, we first compute a global query vector \(\mathbf{Q}_{\text{global}}\) as the average of all query vectors across visual tokens:  \begin{equation}
\mathbf{Q}_{\text{global}} = \frac{1}{n} \sum_{i=1}^{n} \mathbf{Q}_i
\label{globalq}
\end{equation}  

\noindent This vector represents the overall query information requested by visual tokens at layer \(L\) across all attention heads. The importance score for each visual token is then computed by first taking the dot product between its key and the global query for each attention head. A softmax is applied across visual tokens within each attention head, followed by averaging across attention heads. The final score is obtained by scaling the result with the hidden state norm:  

\begin{equation}
\label{equationeuti}
s_i = \frac{1}{n_h} \sum_{j=1}^{n_h} \text{Softmax} \left( \mathbf{k}_i^{(j)} \cdot \mathbf{Q}_{\text{global}}^{(j)} \right) \cdot \|\mathbf{h}_i\|_2
\end{equation}  

\noindent Then, we divide the visual tokens into important and unimportant tokens, using a parameter $\lambda \in [0,1]$ to control the percentage of tokens deemed unimportant. The two sets are defined as follows: \begin{equation}
S_{\text{important}} = \left\{ i \mid s_i \geq \text{Percentile}(s, \lambda) \right\}
\label{important_set}
\end{equation}
\begin{equation}
S_{\text{unimportant}} = \left\{ i \mid s_i < \text{Percentile}(s, \lambda) \right\}
\label{unimportant_set}
\end{equation}
Unimportant tokens can be pruned, or the resulting sets can be combined with a clustering algorithm to further reduce the number of visual tokens, as we will show in the next section. The full \textbf{EUTI} algorithm is illustrated in \cref{algo:pruning}. We note that in the case where Rotary Embeddings are used \cite{rotary}, we use the keys and queries before their application to avoid any positional bias.
\subsection{Clustering-based merging of visual tokens} \label{dbdpc}

\textbf{Distance Bounded Density
Peak Clustering} Relying solely on the importance scores presented above to prune unimportant tokens can lead to a significant reduction in visual tokens, retaining only important ones. However, redundant information may still be present across retained visual tokens. Therefore, we propose merging the redundant visual tokens using a clustering algorithm. We desire our clustering algorithm to have the following characteristics: 

\begin{itemize}[align=left, labelwidth=20pt, itemindent=20pt, labelsep=10pt]
    \item[(a)] Low computational time.
    \item[(b)] Avoid assigning points that are far from each other, \quad \phantom{XXXp} in terms of feature similarity, into the same cluster.
\end{itemize}
\begin{table}[!htbp]
\setlength{\tabcolsep}{2pt}
\centering
\caption{\small \textbf{Throughput ratio, reduction ratio, and GPU memory usage for \textbf{PACT}, FastV, VTW, and ToME applied to LLaVA-OneVision-7B. Results are reported at a 98.6\% Approach-to-Reference Metric Ratio.}}

\label{tab:performance_inference_summary}
    \resizebox{\columnwidth}{!}{
  \begin{tabular}{@{}c c c c c c @{}} 
  
    \toprule
    & \textbf{No reduction} & \textbf{PACT (ours)} & \textbf{FastV} & \textbf{VTW} & \textbf{ToME} \\
    \midrule
     \textbf{Reduction Ratio} & 0\%  & \textbf{71.3}\% & 50\% & 25\% & 40\% \\
    \textbf{LLM Throughput Ratio} & 100\%  & \textbf{225\%} & 165\% & 160\% & 137\% \\
    \textbf{GPU Maximum Memory Consumption (GB)} & 27.4  & \textbf{19.05} & 30.4 & 19.2 & 21.4 \\
    \bottomrule
  \end{tabular}}
\end{table}
\noindent Condition (b) ensures that outliers are not assigned to distant cluster centers, as we speculate that these outliers contain important information and should only be merged with nearby outliers or remain as single points in separate clusters. Condition (b) also guarantees that points in each cluster will be relatively close to each other, which minimizes information loss when assigning a single vector as their representative. The Density Peaks Clustering (DPC) algorithm \cite{dpc} is appealing in this context because it satisfies condition (a), unlike iterative clustering algorithms like k-means \cite{kmeans}. However, DPC does not satisfy condition (b) as it can form large clusters where boundary points may be distant from each other. The same issue arises with other algorithms such as DBSCAN \cite{dbscan}. Therefore, we propose a new clustering algorithm, which we call \textbf{D}istance \textbf{B}ounded \textbf{D}ensity \textbf{P}eaks \textbf{C}lustering (\textbf{DBDPC}).

\noindent \textbf{DBDPC} takes as input a set of vectors \( \{ \mathbf{u}_i \in \mathbb{R}^{d_1} \}_{i=1}^q \), where \( q, d_1 \in \mathbb{N}^+ \), and outputs a set of clusters.
 Our algorithm's output depends on two parameters, the cutoff distance \(d_c \in \mathbb{R}^+\) and a normalization factor \(d_n \in \mathbb{R}^+\), as well as a distance function \(d : \mathbb{R}^{d_1} \times \mathbb{R}^{d_1} \to \mathbb{R}^+\).
 We define the distance between two vectors $\mathbf{u}_i$ and $\mathbf{u}_j$ as:
\begin{equation}
d_{ij} = d(\mathbf{u}_i, \mathbf{u}_j) = 1 - \frac{\mathbf{u}_i \cdot \mathbf{u}_j}{ \| \mathbf{u}_i \|_2 \| \mathbf{u}_j \|_2 }
\label{distance_equation}
\end{equation}
Then the local density $\rho_i$ is calculated as:
\begin{equation}
\rho_i = \sum_{j} e^{-d_{ij}/d_n}
\label{rho_equation}
\end{equation}
\noindent We process the $\mathbf{u}_i$ vectors from highest to lowest $\rho$ values and designate a vector as a cluster center if its minimum distance from already selected centers is greater than $d_c$. Each vector $\mathbf{u}_i$ is then assigned to the cluster of the closest center. Our algorithm guarantees that the distance from each vector to its cluster center is less than \(d_c\), thereby satisfying condition (b) stated above. The full \textbf{DBDPC} algorithm is detailed in \cref{algodbdpc}. The center identification process in \textbf{DBDPC}  ensures that inter-cluster distances are upper-bounded by \( 2d_c \times (2 - d_c) \) while distances between cluster centers are lower-bounded by \( d_c \), which we formally prove in \cref{Dbdpccharac}. We note that several parts of our algorithm are presented as for-loops for clarity. However, all computations are parallelizable on GPU, as there are no dependencies between the elements of each loop, except for the part where we select cluster centers. For this part, we use a recursive algorithm that efficiently identifies an initial set of centers and discarded vectors, thereby reducing the number of vectors to be processed. We explain this in detail in \cref{effecientdbdpc}. For a comparison between \textbf{DBDPC} and DPC, as well as a qualitative comparison with other clustering algorithms, refer to \cref{Qualitative}.

\begin{algorithm}[t!]
\caption{DBDPC}
\label{algodbdpc}
\begin{algorithmic}[0]
\INPUT Cutoff distance \( d_c \in \mathbb{R}^+ \), normalization factor \( d_n \in \mathbb{R}^+ \), set of vectors \( \{ \mathbf{u}_i \in \mathbb{R}^{d_1} \}_{i=1}^q \)
\OUTPUT Cluster center indices \( C_{\text{centers}} \), element indices in each cluster \( C_{\text{elements}} \)

\ForAll{pairs \( (\mathbf{u}_i, \mathbf{u}_j) \)} 
    \State $d_{ij} = 1 - \frac{ \mathbf{u}_i \cdot \mathbf{u}_j }{ \| \mathbf{u}_i \|_2 \| \mathbf{u}_j \|_2 }$
\EndFor

\ForAll{vectors \( \mathbf{u}_i \)} 
    \State $\rho_i = \sum_{j=1}^{q} e^{-d_{ij} / d_n}$
\EndFor

\State Sort vectors by \( \rho_i \) in descending order, obtaining indices \( [i_1, i_2, \dots, i_q] \)
\label{linefor}

\State Initialize $C_{\text{centers}} = \{i_1\}$, $C_{\text{elements}}=\{i_1 : \emptyset\}$
\ForAll{indices \( i_k \) in sorted order} 
    \If{\( \min_{s \in C_{\text{centers}}} d_{i_k s} > d_c \)}
        \State $C_{\text{centers}} = C_{\text{centers}} \cup \{i_k\}$
        \State $C_{\text{elements}}[i_k] = \emptyset$
    \EndIf
\EndFor

\ForAll{indices \( i \)} 
    \State $s_i = argmin_{s \in C_{\text{centers}}} d_{i s}$
    \State $C_{\text{elements}}[s_i] = C_{\text{elements}}[s_i] \cup \{ i \}$
\EndFor

\State \textbf{Return} $C_{\text{centers}}$, $C_{\text{elements}}$
\end{algorithmic}

\end{algorithm}

\noindent \textbf{Which vectors should be used for distance calculation?} As previously discussed, the \textbf{DBDPC} algorithm operates on a set of vectors that are used for distance calculation. To achieve effective clustering, the dot product between these vectors needs to accurately reflect the similarity between the corresponding visual tokens. Fortunately, transformers address this issue through the QKV self-attention mechanism. Specifically, the key vectors \( K \) provide a meaningful representation of each token, tailored for dot product similarity. Therefore, we will use the key vectors in the \textbf{DBDPC} algorithm. Formally, we have:
\begin{equation}
C_{\text{centers}}, \, C_{\text{elements}} = \text{DBDPC}(K')
\label{dbdpc_equation}
\end{equation}
where \( K' = \{ \mathbf{u}_i \in K \mid i \in S_{\text{important}} \} \) is the subset of keys consisting of elements with indices in \( S_{\text{important}} \).

\noindent \textbf{What about unimportant tokens near cluster centers?} Tokens initially deemed unimportant but close enough to cluster centers have a high probability of being mislabeled. We add these tokens to the corresponding cluster to limit information loss. Formally, we define a threshold based on a coefficient \( \alpha \), where any token \( \mathbf{u}_i \), initially excluded, is added to the cluster of the closest center \( s \in C_{\text{centers}} \) if its distance to the center satisfies \( d_{is} < \alpha \cdot d_c \). Specifically, the new cluster elements set \( C_{\text{elements}}^{(s)} \) is updated as follows:
\begin{equation}
\begin{aligned}
S_{\text{added}}^{(s)} &= \left\{ i \in S_{\text{unimportant}} \mid s = argmin_{s' \in C_{\text{centers}}} d_{is'} \, \right. \\
&\hspace{3cm} \left. \text{and  } \, d_{is} < \alpha \cdot d_c \right\}
\end{aligned}
\label{added_set}
\end{equation}

\begin{equation}
C_{\text{elements}}^{(s)} \gets C_{\text{elements}}^{(s)} \cup S_{\text{added}}^{(s)}
\label{update_elements}
\end{equation}

\begin{algorithm}[!t]
\caption{PACT}
\label{pruneandmerge}
\begin{algorithmic}[0]
\INPUT Hidden states $\mathbf{H} = [\mathbf{h}_1, \dots, \mathbf{h}_n] \in \mathbb{R}^{n \times d}$; key and query matrices $\mathbf{K}, \mathbf{Q} \in \mathbb{R}^{n \times n_h \times d_h}$; position IDs $\mathbf{P} = [p_1, \dots, p_n]$; pruning percentage $\lambda \in [0,1]$; cutoff distance $d_c > 0$; tolerance coefficient $\alpha > 0$
\OUTPUT Merged hidden states $\mathbf{H}'$; new position IDs $\mathbf{P}'$
\State \textbf{Step 1: Identify important and unimportant tokens} 
\State $S_{\text{important}},\ S_{\text{unimportant}} \gets \text{EUTI}(\mathbf{H},\ \mathbf{K},\ \mathbf{Q},\ p)$
\State \textbf{Step 2: Cluster important tokens with DBDPC}
\State $\mathbf{K}' \gets \{ \mathbf{k}_i \in \mathbf{K} \mid i \in S_{\text{important}} \}$
\State $C_{\text{centers}},\ C_{\text{elements}} \gets \text{DBDPC}(\mathbf{K}',\ d_c)$
\State \textbf{Step 3: Assign unimportant tokens to sufficiently close clusters.}
\ForAll{$i \in S_{\text{unimportant}}$}
    \State $s_i \gets argmin_{s} d_{is}$
    \If{$d_{is_i} < \alpha.d_c$}
        \State $C_{\text{elements}}^{(s_i)} \gets C_{\text{elements}}^{(s_i)} \cup \{ i \}$
    \EndIf
\EndFor
\State \textbf{Step 4: Merge hidden states and assign position IDs}
\ForAll{$s \in C_{\text{centers}}$}
    \State $\mathbf{h}'_s \gets \frac{1}{|C_{\text{elements}}^{(s)}|} \sum_{i \in C_{\text{elements}}^{(s)}} \mathbf{h}_i$
    \State $p'_s \gets p_s$
\EndFor
\State \textbf{Return} $\mathbf{H}'$, $\mathbf{P}'$
\end{algorithmic}
\end{algorithm}
\noindent \textbf{Merging} Finally, the hidden states corresponding to the elements in each cluster are merged. Formally, the merged hidden states are computed as:
\begin{equation}
\small
\mathbf{H}' = \left\{ \frac{1}{|C_{\text{elements}}^{(j)}|} \sum_{i \in C_{\text{elements}}^{(j)}} \mathbf{h}_i \, \middle| \, C_{\text{elements}}^{(j)} \in C_{\text{elements}} \right\}
\label{H_prime_equation}
\end{equation}

\noindent \textbf{Defining the position IDs} Accurately assigning position IDs to each vector in the new hidden states \(\mathbf{H}'\) is crucial, especially for models using Rotary embeddings, as these IDs determine the input image structure or the temporal dependencies of the input video. In order to achieve a low statistical discrepancy compared to regular inference, we assign the position ID for each vector from $H'$ as its corresponding cluster center. The full \textbf{PACT} pipeline is shown in \cref{pruneandmerge}. When Rotary Embeddings are used, \textbf{DBDPC} uses the keys after these embeddings are applied, whereas \textbf{EUTI} uses the keys and queries before applying these embeddings. For clarity, we omit this detail in \cref{pruneandmerge}. We also note that both \textbf{DBDPC} and \textbf{EUTI}, as well as \textbf{PACT}, do not use textual tokens. Therefore, visual token reduction is performed independently of the textual context, making our method well-suited for multi-turn conversations.
 
\noindent \textbf{Proportional attention} Merging tokens reduces their influence in the attention mechanism and can therefore deteriorate performance if many important tokens are merged together. To mitigate this, we employ proportional attention.
Let \( K \), \( Q \), and \( V \) denote the keys, queries, and values at a layer \( L' \), where \( L' \geq L \). For each attention head \( j \), the attention scores are calculated as follows:
\begin{equation}
A^{(j)} = \text{softmax} \left( \frac{Q^{(j)} K^{(j)\top}}{\sqrt{d_{l'}}} + \log \mathbf{W} + \mathbf{B} \right)
\label{attention_equation}
\end{equation}
where \( d_{l'} \) is the dimensionality of the query for each attention head. Here, \( \mathbf{W} \) is a matrix representing the weight of each token, and \( \mathbf{B} \) is the attention mask. Specifically, for visual tokens, \( w_{i_0,i_1} \) represents the size of the cluster corresponding to token \( i_1 \), for any value of $i_0$. For each textual token at position \( t \), \( w_{i_0,t} = 1 \), as they remain unmerged, retaining a weight of one.
 By scaling the attention scores based on \( \mathbf{W} \), the model effectively treats each visual token as if it represents multiple tokens. We note that when using proportional attention, we use PyTorch's scaled dot-product attention, which produces similar results to the official FlashAttention implementation while supporting custom masks.
\\
\textbf{Selecting the layer $L$ for token reduction:}\label{choosinglayer} To ensure maximum computational gain, we must choose an early layer \(L\) for visual token reduction. However, we also require that the keys at the selected layer are not too similar, allowing for effective clustering and pruning. Thus, we select the earliest layer where the maximum distance between keys is sufficiently high. \Cref{plotdistancekeys} shows that in the initial layers of LLaVA-OneVision-7B, the keys corresponding to visual tokens are quite similar, indicating a lack of distinctive features necessary for effective pruning and clustering.
\begin{figure}[!t]
    \centering

    \begin{minipage}[t]{0.235\textwidth}
        \centering
        \includegraphics[width=\textwidth]{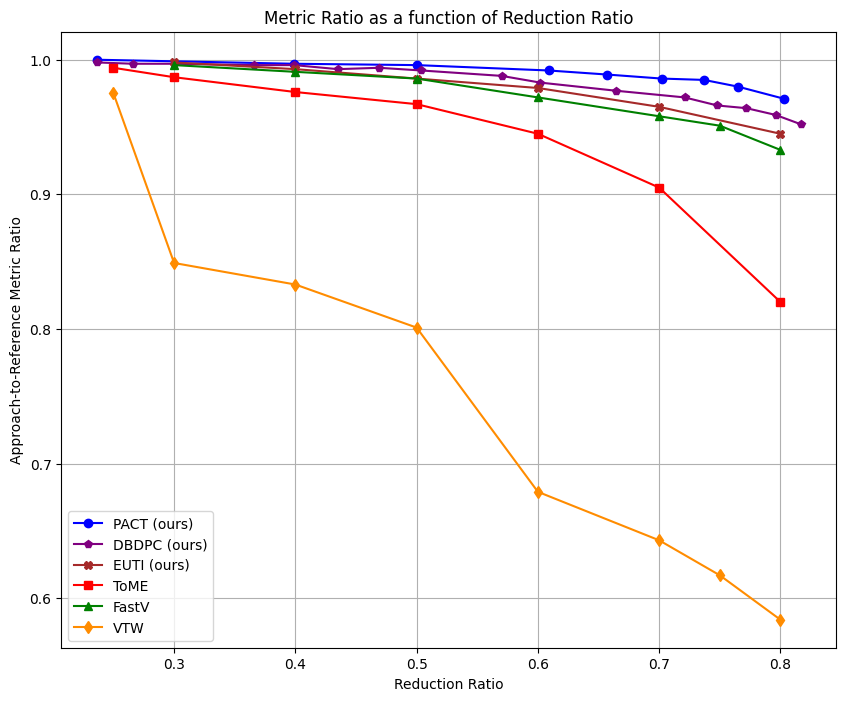}
        \label{fig:prunenmerge_vs_others_red}
    \end{minipage}
    \hfill
    \begin{minipage}[t]{0.235\textwidth}
        \centering
        \includegraphics[width=\textwidth]{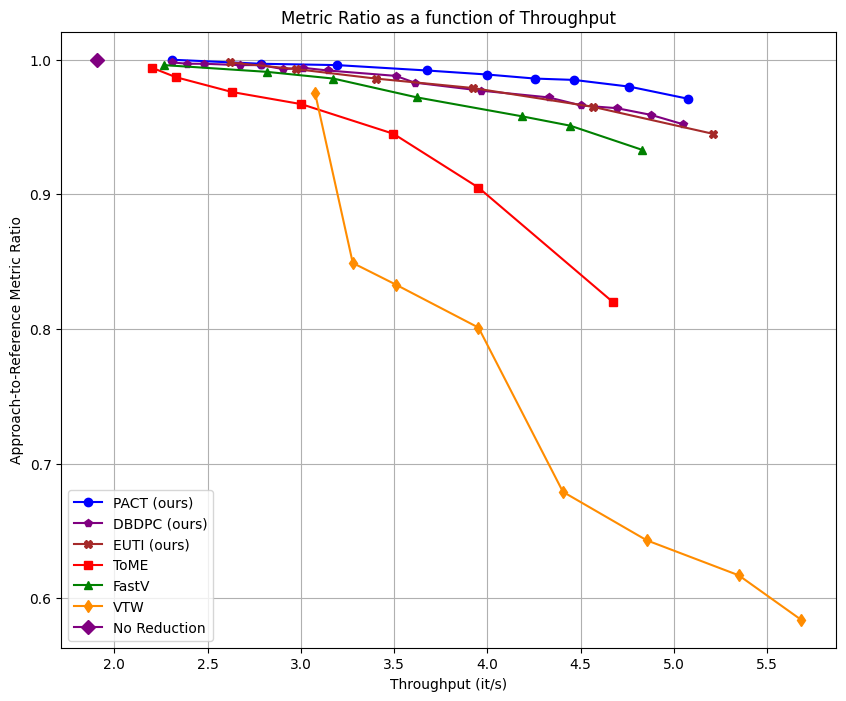}
        \label{fig:prunenmerge_vs_others_Throughput}
    \end{minipage}
    \caption{\textbf{Comparison between \textbf{PACT}, \textbf{DBDPC}, and \textbf{EUTI} against other visual token reduction methods across various reduction ratios applied on LLaVA-OneVision-7B.}}
    \label{fig:prunenmerge_vs_others}
\end{figure}



\section{Experiments}

\subsection{Evaluation datasets}

We evaluate the effectiveness of \textbf{PACT} using diverse benchmarks, similar to those used for LLaVA-OneVision-7B, covering single-image, multi-image, and video tasks. We use AI2D \cite{AI2D}, TextVQA \cite{TextVQA}, ChartQA \cite{ChartQA}, DocVQA \cite{DocVQA}, and InfographicVQA \cite{Infographicvqa} to assess \textbf{PACT}'s ability to reduce visual tokens while maintaining performance in text-rich documents. To test reasoning across multiple disciplines, we use MME \cite{MME}, MMBench \cite{MMBench}, MMVet \cite{MMVet}, MathVerse \cite{MathVerse}, MathVista \cite{MathVista}, MMMU \cite{MMMU}, MMStar \cite{MMStar}, and ScienceQA \cite{ScienceQA}. Additionally, Vibe-Eval \cite{Vibe-Eval}, MM-LiveBench \cite{LMMs-Eval}, and LLaVA-Bench-Wilder \cite{lavanext} evaluate its robustness in real-world scenarios and visual chat contexts. We use LLaVA-Interleave Bench \cite{lavanext} and MuirBench \cite{MuirBench} to examine \textbf{PACT}'s efficiency in token reduction while preserving inter-image reasoning.  To assess performance in video comprehension tasks, we use ActivityNet-QA \cite{ActivityNet-QA}, MLVU \cite{MLVU}, VideoMME \cite{VideoMME}, EgoSchema \cite{Egoschema}, and PerceptionTest \cite{PerceptionTest}. Finally, Video-ChatGPT \cite{Video-ChatGPT} evaluates the method's effectiveness in dialogue-based video interaction.

\begin{figure}[!t]
    \centering

    \begin{minipage}[t]{0.235\textwidth}
        \centering
        \includegraphics[width=\textwidth]{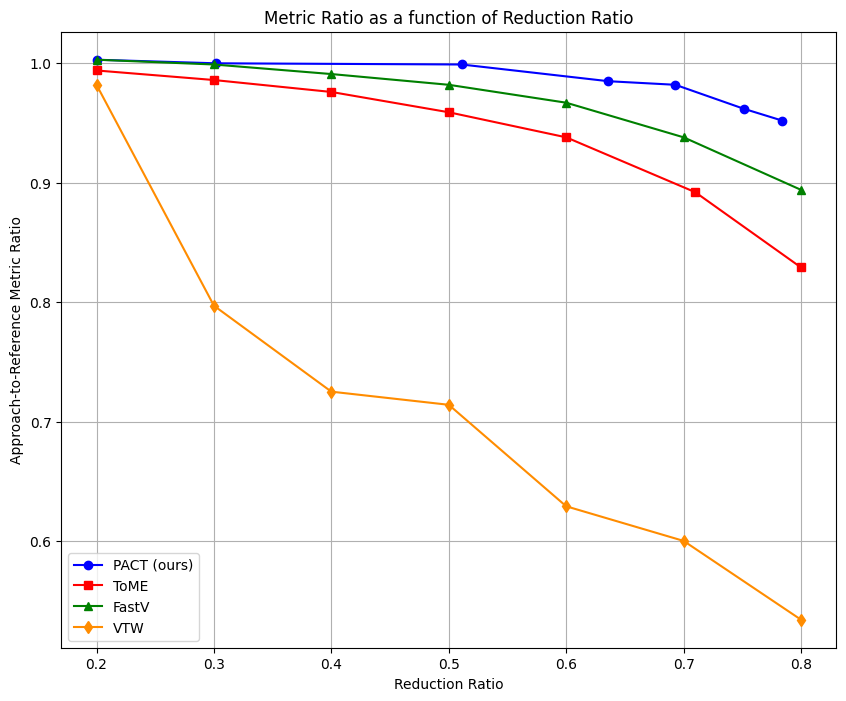}
    \end{minipage}
    \hfill
    \begin{minipage}[t]{0.235\textwidth}
        \centering
        \includegraphics[width=\textwidth]{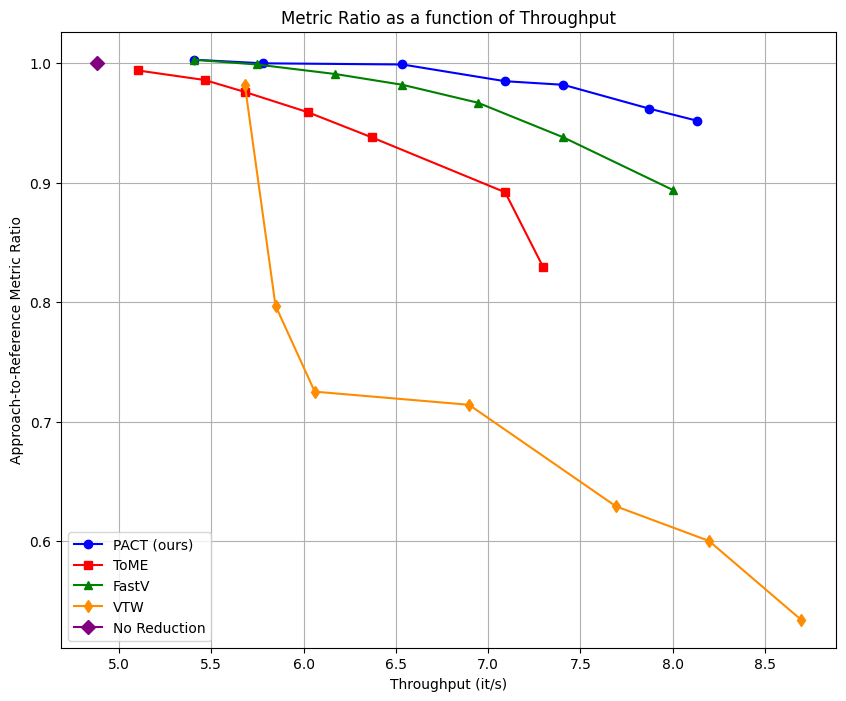}
    \end{minipage}
    \caption{\textbf{Comparison between \textbf{PACT}  and other visual token reduction methods across various reduction ratios applied on Qwen2-VL-7B-Instruct.}}
    \label{figresutsqwen}
\end{figure}

\begin{figure}[!b]
    \centering

    \begin{minipage}[t]{0.235\textwidth}
        \centering
        \includegraphics[width=\textwidth]{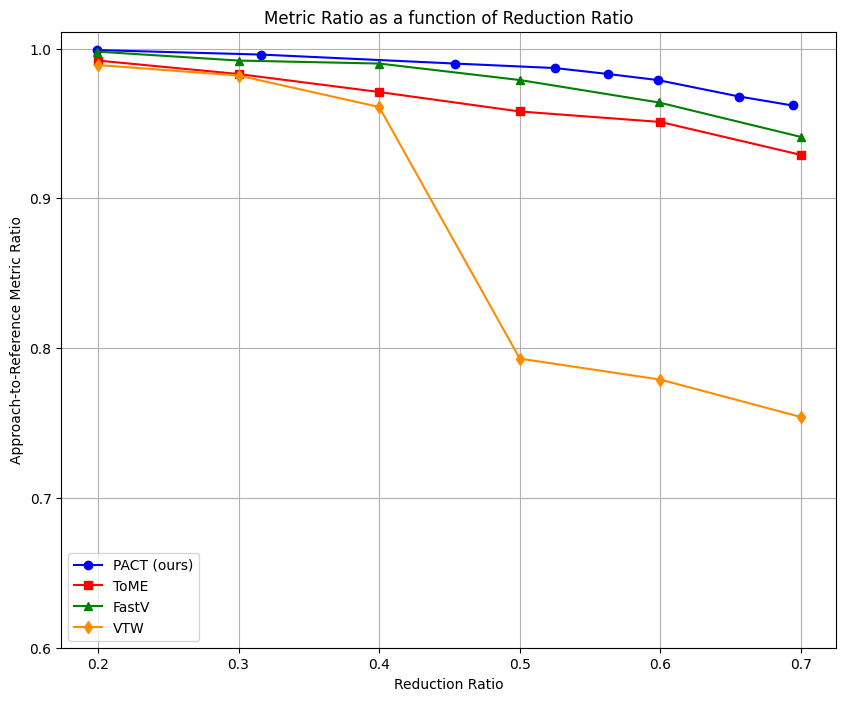}
    \end{minipage}
    \hfill
    \begin{minipage}[t]{0.235\textwidth}
        \centering
        \includegraphics[width=\textwidth]{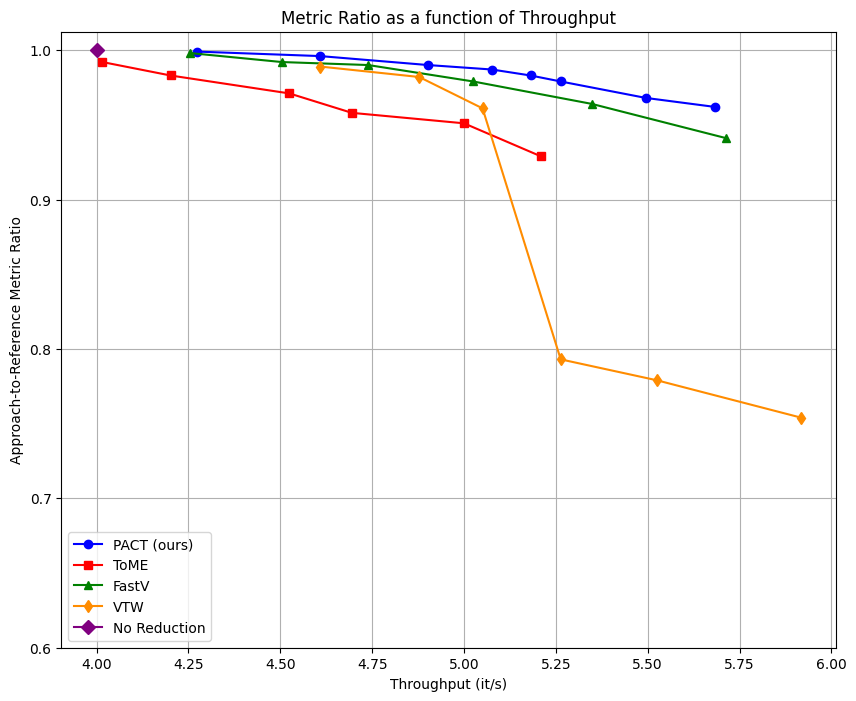}
    \end{minipage}
  \caption{\textbf{Comparison between \textbf{PACT}  and other visual token reduction methods across various reduction ratios applied on InternVL2-8B.}}
    \label{figresutsintern}
\end{figure}

\subsection{Evaluation setup}
\begin{table*}[!ht]
\centering
\caption{\small \textbf{Comparison of \textbf{PACT} with FastV, VTW, and ToME on LLaVA-OneVision-7B.} \textbf{Algo. Time} refers to the average time the algorithm takes per input element, measured in seconds. \textbf{Proc. Time} refers to the average time taken by both the language model and the reduction algorithm per input element.  \textbf{Red. Ratio} stands for average Reduction Ratio. The Algo. Time for VTW is nearly zero, and thus omitted. The different visual token reduction methods are evaluated at the same reduction ratio as \textbf{PACT}.}
\label{tab:comparison_results}
\resizebox{\textwidth}{!}{
  \begin{tabular}{@{} c c c c c c c c c c c c c c c @{}} 
    \toprule
\textbf{Dataset} & \multicolumn{2}{c}{{\textbf{No reduction}}} & \multicolumn{4}{c}{\textbf{PACT (Ours)}} & \multicolumn{3}{c}{\textbf{FastV}} & \multicolumn{2}{c}{\textbf{VTW}} & \multicolumn{3}{c}{\textbf{ToME}} \\
    \cmidrule(lr){2-3} \cmidrule(lr){4-7} \cmidrule(lr){8-10} \cmidrule(lr){11-12} \cmidrule(lr){13-15}
 & \textbf{Metric} & \textbf{Proc. Time} & \textbf{Metric} & \textbf{Red. Ratio} & \textbf{Proc. Time} & \textbf{Algo. Time} & \textbf{Metric} & \textbf{Proc. Time} & \textbf{Algo. Time} & \textbf{Metric} & \textbf{Proc. Time} & \textbf{Metric} & \textbf{Proc. Time} & \textbf{Algo. Time} \\
    \midrule
\textbf{VideoMME} & 58.5 & 0.792 & \textbf{57.6} & 65.6\% & 0.369 & 0.021 & 57.0 &  0.371 & 0.040 & 46.9 & 0.296 & 57.0 & 0.417 &  0.091 \\
\textbf{MME} & 1579 & 0.554 & 1564.0 & 70.2\% & 0.243 & 0.017 & \textbf{1576.0} & 0.244 & 0.016 &  842.0 & 0.231 & 1556.9 & 0.317 & 0.084 \\
\textbf{DocVQA} & 87.2 & 1.088 & \textbf{84.4} & 67.9\% & 0.519 & 0.026 & 84.3 & 0.524 & 0.051 & 10.5 & 0.449 & 61.9 & 0.576 &  0.099 \\
\textbf{MLVU} & 65.2 & 0.795 & \textbf{64.7} & 66.4\% & 0.361 & 0.022 & 62.9 & 0.369 & 0.040 & 54.4 & 0.312 & 63.4 & 0.417 &  0.092 \\
\textbf{LLaVA-Interleave} & 64.1 & 0.249 & \textbf{64.0} & 69.7\% & 0.133 & 0.010 & 58.9 & 0.139 & 0.007 & 32.4 & 0.123 & 50.3 & 0.192 & 0.068 \\
\textbf{ChartQA} & 79.9 & 0.671 & 76.5 & 68.5\% & 0.341 & 0.019 & \textbf{77.0} & 0.342 & 0.016 & 16.6 & 0.307 & 63.4 & 0.402 & 0.082 \\
\textbf{MMBench} & 80.6 & 0.249 & \textbf{80.3} & 69.3\% & 0.135 & 0.010 & 79.0 & 0.140 & 0.005 & 52.4 & 0.125 & 79.7 & 0.193 & 0.066 \\
\textbf{MuirBench} & 42.0 & 0.384 & \textbf{43.1} & 67.8\% & 0.178 & 0.013 & 40.4 & 0.178 & 0.009 & 34.9 & 0.162 & 40.5 & 0.233 & 0.072 \\
\textbf{ScienceQA} & 95.9 & 0.238 & \textbf{93.8} & 69.6\% & 0.133 & 0.010 & 91.6 & 0.137 & 0.006 & 80.0 & 0.124 & \textbf{93.8} & 0.190 & 0.066 \\
\textbf{MMMU} & 49.2 & 0.139 & \textbf{48.9} & 70.4\% & 0.104 & 0.007 & \textbf{48.9} & 0.106 & 0.003 & 43.5 & 0.093 & 48.6 & 0.124 & 0.062 \\
\textbf{AI2D} & 81.5 & 0.382 & \textbf{81.0} & 69.8\% & 0.186 & 0.013 & 79.4 & 0.191 & 0.014 &  69.7 & 0.177 & 79.7 & 0.244 & 0.073 \\
\textbf{InfographicVQA} & 66.0 & 0.895 & \textbf{61.9} & 64.7\% & 0.481 & 0.023 & 58.6 & 0.483 & 0.040 & 24.5 & 0.408 &  48.3 & 0.607 & 0.130 \\
\textbf{MMStar} & 62.0 & 0.297 & \textbf{60.1} & 69.7\% & 0.147 & 0.011 & 58.6 & 0.152 & 0.007 & 37.2 & 0.165 & \textbf{60.1} & 0.229 & 0.069 \\
\textbf{ActivityNetQA} & 54.5 & 0.921 & \textbf{55.1} & 70.0\% & 0.419 & 0.029 & 53.7 & 0.425 & 0.042 & 36.6 & 0.394 & 54.1 & 0.513 & 0.203 \\
\textbf{MM-LiveBench} & 73.1 & 4.434 & \textbf{71.7} & 67.5\% & 3.212 & 0.047 & 64.4 & 3.221 & 0.044 & 41.0 & 3.080 & 64.2 & 3.607 & 0.102 \\
\textbf{LLaVA-Wilder} & 71.0 & 10.10 & \textbf{71.5} & 70.0\% & 8.262 & 0.035 & 71.0 & 8.263 & 0.025 & 48.8 & 7.515 & 68.0 & 7.926 & 0.085 \\
\textbf{MathVerse} & 16.8 & 0.831 & 16.6 & 74.2\% & 0.361 & 0.021 & 16.1 & 0.382 & 0.036 & \textbf{17.6} & 0.301 & 16.5 & 0.559 & 0.150 \\
\textbf{MathVista} & 63.3 & 0.440 & \textbf{62.0} & 70.7\% & 0.271 & 0.015 & 59.5 & 0.272 & 0.016 & 38.5 & 0.260 & 55.0 & 0.338 & 0.071 \\
\textbf{MMVet} & 58.0 & 4.602 & \textbf{58.4} & 70.4\% & 3.793 & 0.035 & 51.7 & 3.795 & 0.036 & 15.7 & 3.652 & 47.2 & 4.115 & 0.212 \\
\textbf{Vibe-Eval} & 41.6 & 5.153 & \textbf{39.1} & 71.1\% & 3.709 & 0.032 & 38.2 & 3.714 & 0.047 & 12.3 & 3.550 & 31.2 & 4.317 & 0.095 \\
\textbf{VideoChatGPT} & 3.25 & 2.972 & \textbf{3.25} & 67.2\% & 1.863 & 0.029 & 3.22 & 1.866 & 0.040 & 1.92 & 1.320 & 3.19 & 1.975 & 0.205 \\
\textbf{EgoSchema} & 60.1 & 0.811 & \textbf{60.1} & 66.6\% & 0.351 & 0.021 & 58.7 & 0.353 & 0.044 & 44.8 & 0.297 & 59.8 & 0.391 & 0.091 \\
\textbf{PerceptionTest} & 52.1 & 0.801 & \textbf{52.3} & 66.9\% & 0.353 & 0.023 & 51.7 & 0.357 & 0.040 & 45.0 & 0.296 & 51.1 & 0.393 & 0.090 \\
\textbf{TextVQA} & 75.8 & 0.690 & 75.0 & 67.2\% & 0.332 & 0.023 & \textbf{75.5} & 0.336 & 0.029 & 11.6 & 0.287 & 62.5 & 0.392 & 0.087 \\
    \bottomrule
  \end{tabular}}
\end{table*}
    In our comparison, we include approaches where the reduction is applied at a single layer, similar to \textbf{PACT}, such as FastV and clustering-based visual token reduction. For these approaches, we refer to the reduction ratio as the relative reduction in the number of visual tokens, defined as \(1 - \frac{\text{number of visual tokens after reduction}}{\text{number of visual tokens before reduction}}\). For all these approaches, we use the same value of $L$ and vary hyperparameters to test across different reduction ratios. For methods that use progressive token reduction, like ToME \cite{tome}, or apply reduction after the visual encoder, as PruMerge and HiReD, or when the reduction ratio cannot be controlled at a fixed layer, such as VTW, we adjust the parameters of these approaches to achieve the same average number of visual tokens across all layers as the one-layer reduction methods for a given reduction ratio. When evaluating clustering algorithms for visual token reduction, we apply proportional attention, as it consistently improves performance across all clustering algorithms, especially at high reduction ratios. Additionally, it is crucial to correctly assign position IDs to the resulting reduced set of visual tokens. Details on the assignment strategy are presented in \cref{clusteringpositionalid}. When reporting processing time or throughput, we take into account the total time required by both the language model and the reduction algorithm per input element. In the next section, we base our comparison on a metric called the Approach-to-Reference Metric Ratio, defined as the average of the ratio of the metric of the tested approach to the metric obtained without visual token reduction across all test datasets. Formally we have \textit{Approach-to-Reference Metric Ratio} $ =\frac{1}{N} \sum_{i=1}^{N} \frac{\text{Metric}_{\text{with reduction}}(i)}{\text{Metric}_{\text{no reduction}}(i)}$ where \( N \) is the total number of test datasets. This metric indicates how much of the original model capacity is retained. It is important to note that when using ToME for visual token reduction, a reduction ratio greater than 50\% can't be achieved if the number of visual tokens is reduced by a fixed amount in each layer, as suggested in \cite{tome}. Instead, we use a scheduler to achieve higher reduction ratios, which we explain in \cref{moreabouttome}. More details on the hyperparameters used for evaluating \textbf{PACT} are provided in \cref{implentationdetails}. We follow the same dataset splits and metrics used for evaluating LLaVA-OneVision wherever feasible. More details are provided in \cref{datasetsapendix}. Note that all experiments were conducted on a single A100 GPU.

\begin{figure}[!b]
    \centering

    \begin{minipage}[t]{0.235\textwidth}
        \centering
        \includegraphics[width=\textwidth]{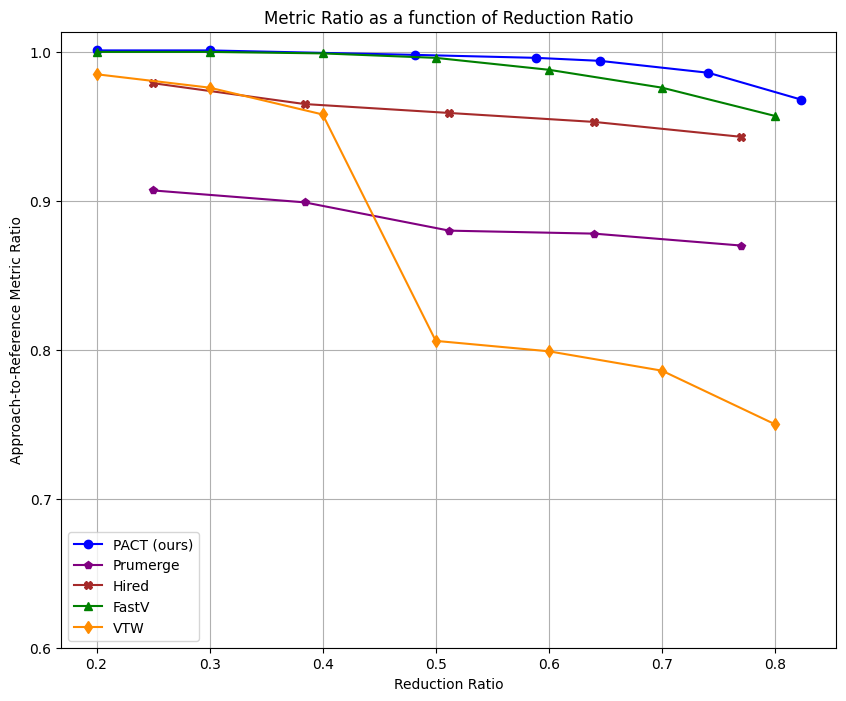}
    \end{minipage}
    \hfill
    \begin{minipage}[t]{0.235\textwidth}
        \centering
        \includegraphics[width=\textwidth]{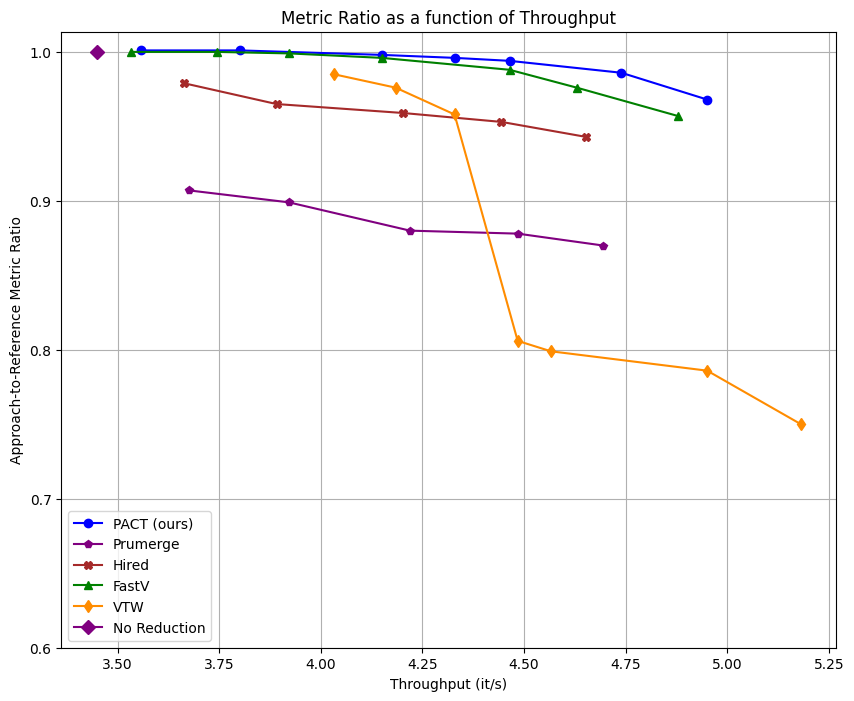}
    \end{minipage}
  \caption{\textbf{Comparison between \textbf{PACT}  and other visual token reduction methods across various reduction ratios applied on LLaVA-1.6-Mistral-7B.}}
    \label{fig:next}
\end{figure} 

\subsection{Results}

\begin{figure}[!b]
    \centering

    \begin{minipage}[t]{0.235\textwidth}
        \centering
        \includegraphics[width=\textwidth]{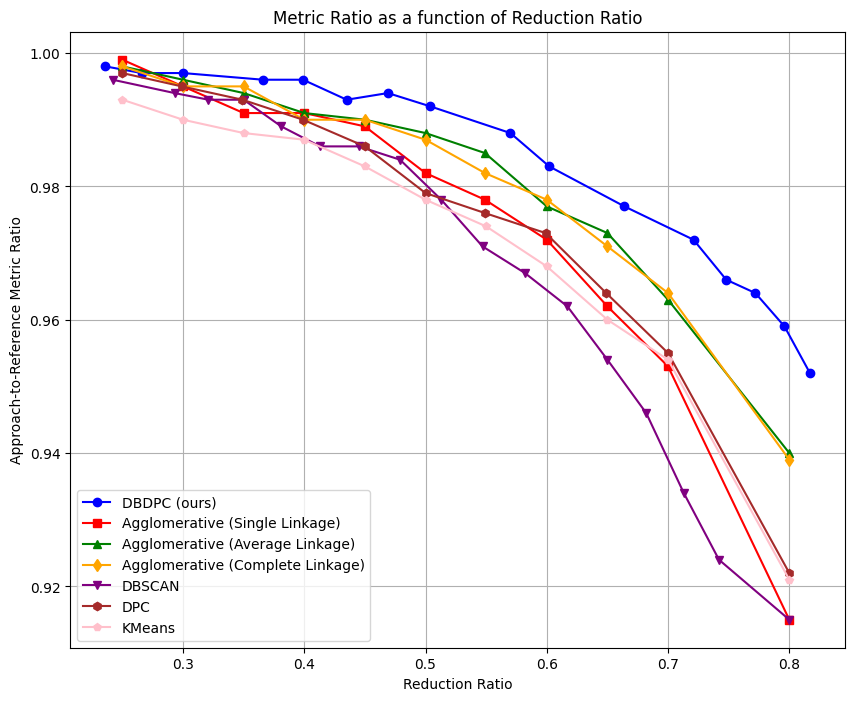}
        \label{fig:DBDPC_vs_others_red}
    \end{minipage}
    \hfill
    \begin{minipage}[t]{0.235\textwidth}
        \centering
        \includegraphics[width=\textwidth]{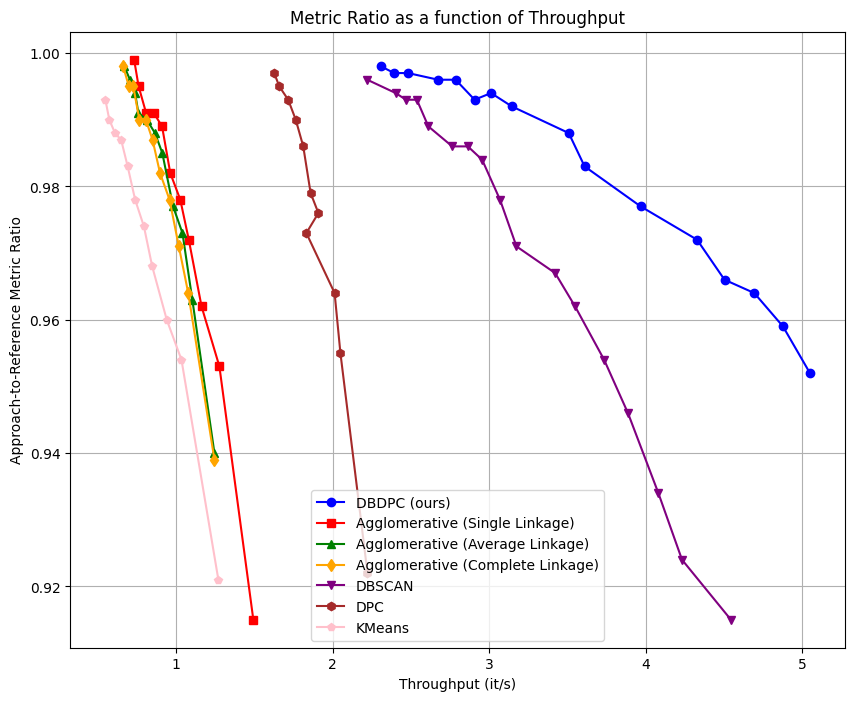}
        \label{fig:DBDPC_vs_others_Throughput}
    \end{minipage}
    \caption{\textbf{Comparison of \textbf{DBDPC} and other clustering algorithms for visual token reduction at different reduction ratios on LLaVA-OneVision-7B.}}
    \label{fig:DBDPC_vs_others}
\end{figure} 

We compare \textbf{PACT} with FastV \cite{fastv}, VTW \cite{tokenwith}, ToME \cite{tome}, PruMerge \cite{llavapurge} and HiRED \cite{hired} on LLaVA-OneVision-7B, InternVL2-8B, Qwen2-VL-7B-Instruct and LLaVA-1.6-Mistral-7B. Since HiRED and PruMerge are only applicable to LLaVA-1.6, we exclude them from other comparisons. As shown in figures \ref{fig:prunenmerge_vs_others}, \ref{figresutsqwen}, \ref{figresutsintern}, and \ref{fig:next} \textbf{PACT} consistently outperforms other methods at both equal reduction ratios and equal throughput across all four models. VTW experiences a significant performance drop for reduction ratios above 40\%, indicating that removing all visual tokens is only effective when done in later layers. FastV and ToME struggle at high reduction ratios, while PruMerge and HiRED exhibit degradation even at low reduction ratios. Meanwhile, \textbf{PACT} maintains acceptable performance even at high reduction ratios. \Cref{tab:comparison_results} and \Cref{resultsqwen} shows that \textbf{PACT} outperforms other approaches on most of the test datasets when applied on LLaVA-OneVision-7B and Qwen2-VL-7B-Instruct. The same conclusion applies to other models, with detailed results provided in \cref{additionalnumericalresults}. In \cref{tab:performance_inference_summary}, we report the reduction ratio, throughput, and maximum GPU memory consumption of the different approaches at an equal Approach-to-Reference Metric Ratio of 98.6\% on LLaVA-OneVision-7B. \textbf{PACT} significantly outperforms the other methods, achieving a reduction ratio of 71.3\%, a GPU memory reduction of 31\%, and a 225\% speedup in the language model's inference time. The per-dataset results used to compute these metrics are shown in \cref{tab:comparison_results_pact_only}. \cref{tab:performance_inference_summary} also indicates that when using FastV, the maximum GPU memory consumption is relatively high due to the costly computation of attention scores. We further compare \textbf{DBDPC} against agglomerative clustering \cite{agglomerative}, k-means \cite{kmeans}, Density Peaks Clustering (DPC) \cite{dpc}, and DBSCAN \cite{dbscan}, with results presented in \cref{fig:DBDPC_vs_others}. The graphs reveal that \textbf{DBDPC} consistently outperforms other clustering algorithms for visual token reduction, exhibiting less performance degradation at equal reduction ratios and demonstrating improved computational efficiency, leading to better throughput. These results validate our hypothesis that, for an effective visual token reduction, it is necessary to ensure that the distances between elements within each cluster do not exceed a predefined threshold. \cref{fig:prunenmerge_vs_others} also shows that \textbf{EUTI} consistently outperforms FastV at equal reduction ratios and is less costly, as it does not require the computation of attention scores. In addition, unlike FastV, \textbf{EUTI} does not introduce a GPU memory overhead\footnote{EUTI achieves roughly the same memory reduction as PACT.}. We provide additional numerical results in \cref{additionalnumericalresults}.

\begin{table*}[!ht]
\centering
\caption{\small \textbf{Comparison of \textbf{PACT} with FastV, VTW, and ToME applied on Qwen2-VL-7B-Instruct across Various Datasets.}}
\label{resultsqwen}
\resizebox{\textwidth}{!}{
  \begin{tabular}{@{} c c c c c c c c c c c c c @{}} 
    \toprule
\textbf{Dataset} & \multicolumn{2}{c}{{\textbf{No Reduction}}} & \multicolumn{3}{c}{\textbf{PACT (Ours)}} & \multicolumn{2}{c}{\textbf{FastV}} & \multicolumn{2}{c}{\textbf{VTW}} & \multicolumn{2}{c}{\textbf{ToME}} \\
    \cmidrule(lr){2-3} \cmidrule(lr){4-6} \cmidrule(lr){7-8} \cmidrule(lr){9-10} \cmidrule(lr){11-12}
 & \textbf{Metric} & \textbf{Proc. Time} & \textbf{Metric} & \textbf{Red. Ratio} & \textbf{Proc. Time} & \textbf{Metric} & \textbf{Proc. Time} & \textbf{Metric} & \textbf{Proc. Time} & \textbf{Metric} & \textbf{Proc. Time} \\
    \midrule
\textbf{MME} & 1654.5 & 0.238 & \textbf{1666.5} & 86.3\% & 0.110 & 1500.0 & 0.111 & 709.24 & 0.120 & 1610.9 & 0.140 \\
\textbf{DocVQA} & 93.9 & 0.516 & \textbf{90.5} & 77.5\% & 0.294 & 86.6 & 0.298 & 8.5 & 0.249 & 42.9 & 0.350 \\
\textbf{TextVQA} & 81.8 & 0.155 & \textbf{80.4} & 67.5\% & 0.132 & 79.9 & 0.135 & 13.2 & 0.118 & 66.2 & 0.151 \\
\textbf{InfographicVQA} & 74.6 & 0.478 & \textbf{70.6} & 69.7\% & 0.278 & 63.3 & 0.273 & 21.5 & 0.225 & 43.9 & 0.299 \\
\textbf{ChartQA} & 80.8 & 0.145 & \textbf{76.0} & 61.1\% & 0.135 & 69.2 & 0.134 & 12.9 & 0.123 & 55.1 & 0.155 \\
\textbf{MMBench} & 77.6 & 0.074 & \textbf{77.1} & 51.5\% & 0.077 & \textbf{77.1} & 0.074 & 76.9 & 0.073 & 75.9 & 0.080 \\
\textbf{MuirBench} & 40.7 & 0.159 & \textbf{41.2} & 76.9\% & 0.113 & 40.4 & 0.112 & 37.9 & 0.111 & 75.8 & 0.125 \\
\textbf{MMMU} & 51.4 & 0.109 & \textbf{51.2} & 72.6\% & 0.093 & 49.3 & 0.092 & 45.4 & 0.088 & 48.9 & 0.105 \\
\textbf{AI2D} & 79.9 & 0.105 & \textbf{78.4} & 64.2\% & 0.096 & 76.2 & 0.097 & 69.0 & 0.087 & 76.4 & 0.115 \\
\textbf{MMStar} & 56.0 & 0.072 & \textbf{54.8} & 61.3\% & 0.072 & 51.5 & 0.067 & 40.3 & 0.065 & 53.8 & 0.077 \\
\textbf{EgoSchema} & 62.1 & 0.360 & \textbf{61.6} & 60.0\% & 0.207 & 60.2 & 0.212 & 46.3 & 0.190 & 61.2 & 0.230 \\
\textbf{MathVerse} & 25.3 & 0.620 & \textbf{24.5} & 82.2\% & 0.393 & 23.7 & 0.396 & 13.9 & 0.296 & 18.1 & 0.651 \\
\textbf{MathVista} & 59.2 & 0.249 & \textbf{57.7} & 73.3\% & 0.195 & 56.4 & 0.194 & 36.8 & 0.165 & 53.5 & 0.275 \\
\textbf{MMVet} & 24.9 & 4.700 & \textbf{25.1} & 80.3\% & 3.820 & 22.3 & 3.830 & 2.7 & 3.650 & 16.7 & 4.780 \\
\textbf{Vibe-Eval} & 47.5 & 3.200 & \textbf{46.1} & 85.0\% & 2.310 & 44.3 & 2.375 & 13.1 & 1.993 & 29.6 & 3.620 \\
\textbf{LLaVA-Interleave} & 35.9 & 0.120 & \textbf{35.5} & 73.7\% & 0.100 & 34.7 & 0.101 & 33.2 & 0.096 & 35.3 & 0.125 \\
\textbf{MM-LiveBench} & 72.6 & 3.970 & \textbf{70.7} & 77.1\% & 3.040 & 63.0 & 3.120 & 39.7 & 2.970 & 57.6 & 4.450 \\
    \bottomrule
  \end{tabular}}
 \end{table*}

\subsection{Ablation study}
\begin{figure}[!b]
    \centering

    \begin{minipage}[t]{0.235\textwidth}
        \centering
        \includegraphics[width=\textwidth]{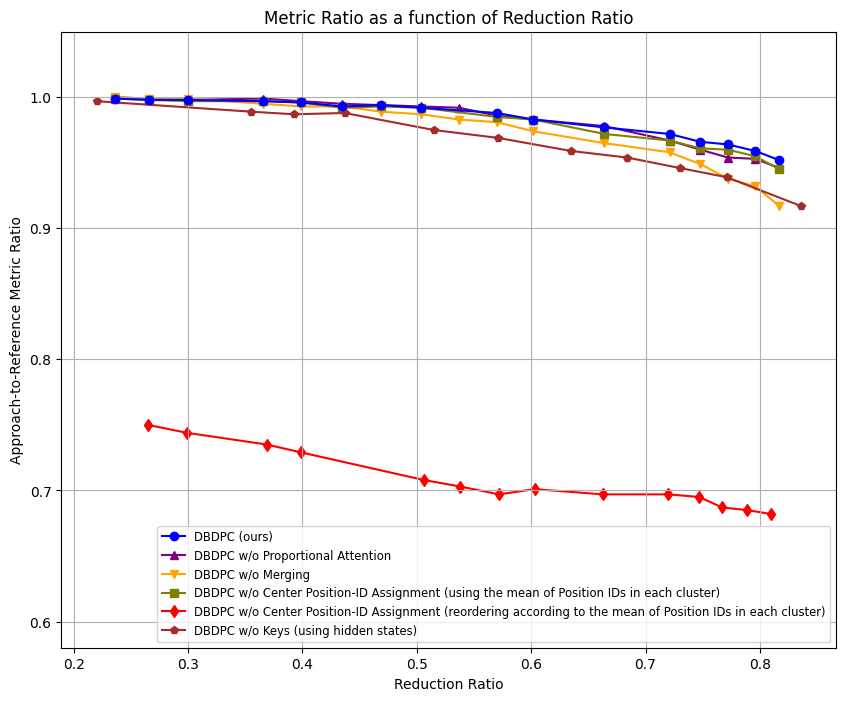}
    \end{minipage}
    \hfill
    \begin{minipage}[t]{0.235\textwidth}
        \centering
        \includegraphics[width=\textwidth]{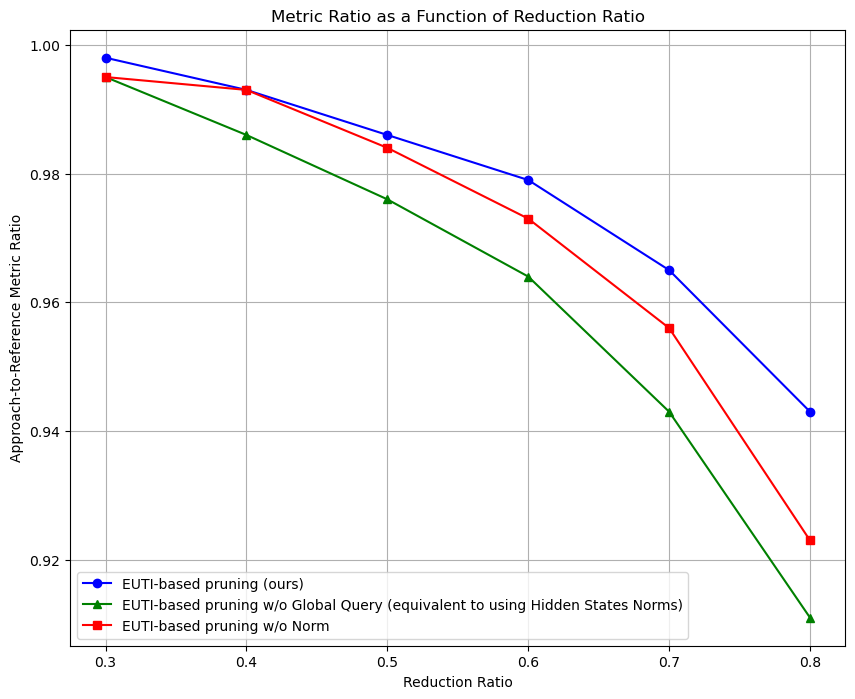}
    \end{minipage}
\caption{\textbf{Ablation study of \textbf{DBDPC} and \textbf{EUTI} on LLaVA-OneVision-7B.}}
    \label{fig:ablation}
\end{figure}

\begin{figure}[!b]
    \centering
    \vspace{-5pt}
    
    \begin{minipage}[t]{0.215\textwidth}
        \centering
        \includegraphics[width=\textwidth]{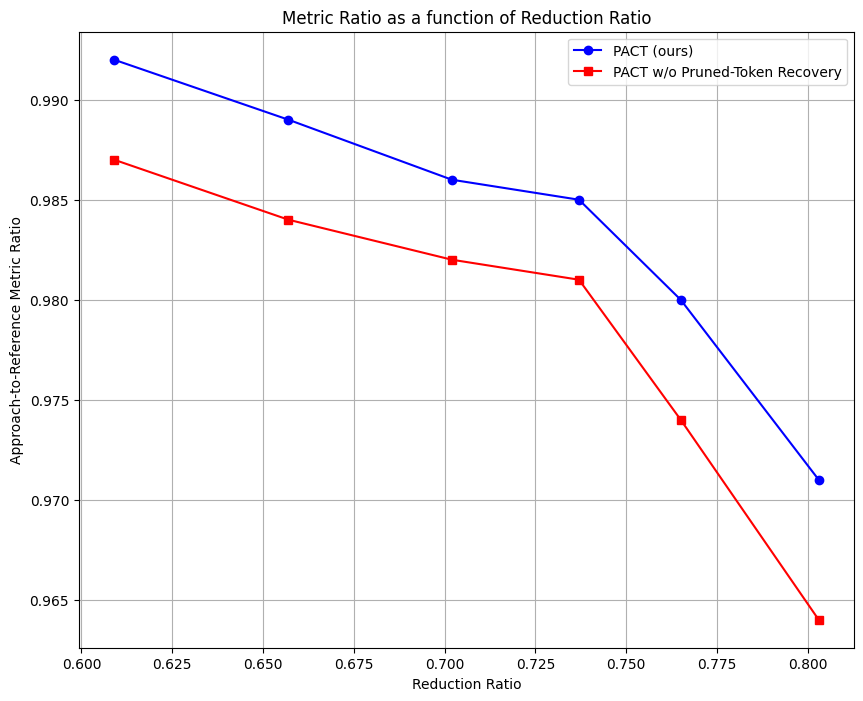}
    \end{minipage}
    \hfill
    \begin{minipage}[t]{0.255\textwidth}
        \centering
        \includegraphics[width=\textwidth]{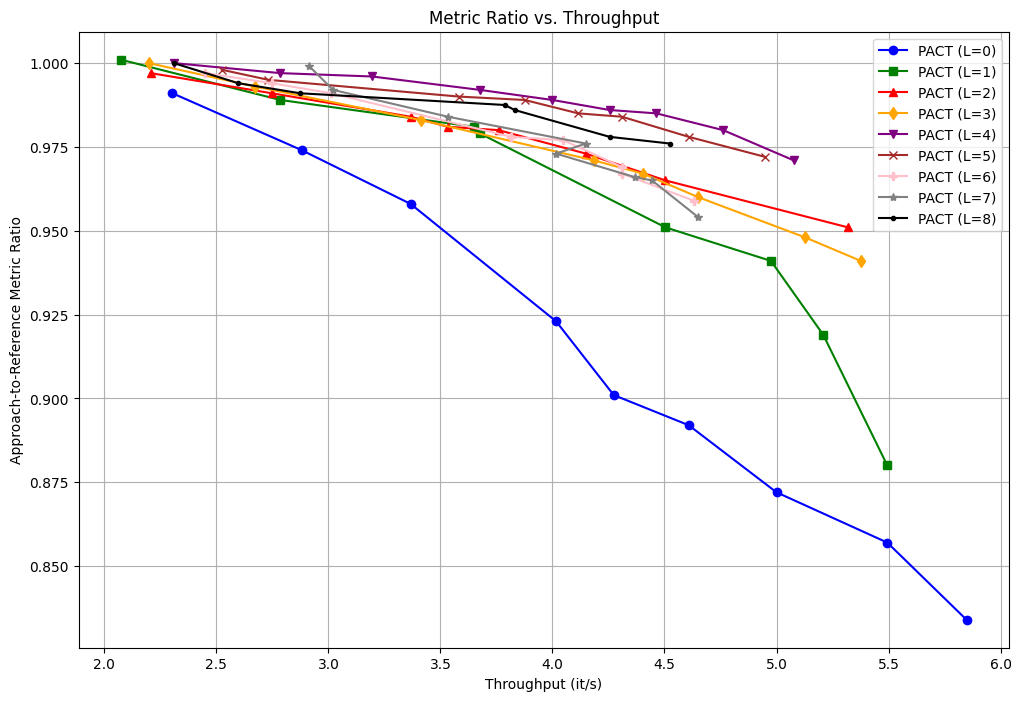}
    \end{minipage}
   \caption{\textbf{Ablation study of \textbf{PACT} on LLaVA-OneVision-7B.} }
    \label{fig:prunemergeablation}
\end{figure}

\label{ablation}\cref{fig:prunenmerge_vs_others} shows that \textbf{PACT} consistently outperforms both \textbf{DBDPC} and \textbf{EUTI} across various reduction ratios. This confirms that combining clustering and pruning techniques yields better performance than using each approach independently, as the combined method addresses both visual tokens irrelevance and redundancy. We ablate several components of the \textbf{DBDPC} algorithm and present the results in \cref{fig:ablation}. First, we ablate token merging by selecting the center of each cluster as the representative token instead of merging tokens within each cluster. We also ablate the use of proportional attention. Additionally, we ablate the assignment of position IDs to the reduced set of tokens and experiment with two alternatives: using the mean of position IDs of all elements in each cluster and assigning position IDs sequentially after reordering the reduced set according to the mean of position IDs. Finally, we ablate the use of key vectors in the clustering process and instead use hidden states. Our results show that each ablated component contributes positively to the performance of the \textbf{DBDPC} algorithm. Notably, correctly assigning position IDs to the reduced set is crucial, as these position IDs reflect the structure of input images and the temporal order of input videos. Additionally, proportional attention proves effective at higher reduction ratios, while token merging enhances performance once the reduction ratio exceeds 50\%. The figure also confirms that keys are better suited for cosine similarity-based distance calculations, as they are naturally used in dot products within the attention mechanism. We perform two separate ablations on \cref{equationeuti} of the \textbf{EUTI} algorithm. The first ablation removes the use of hidden state norms, while the second ablates the use of the global query, which corresponds to using only the hidden state norms. The results in \cref{fig:ablation} show that combining both the global query-based score and the norm of hidden states consistently leads to better results than using either metric alone, suggesting that they provide complementary information about the importance of each visual token. Finally, we ablate the pruned token recovery module in \textbf{PACT} by setting \(\alpha\) to zero, with results presented in \cref{fig:prunemergeablation}. The plot shows that reintegrating visual tokens initially deemed unimportant but close enough to a cluster center consistently enhances performance across different reduction ratios, supporting our hypothesis that these tokens were likely mislabeled by the \textbf{EUTI} module. \Cref{fig:prunemergeablation} also shows the effect of the choice of the reduction layer on \textbf{PACT}'s performance, demonstrating the effectiveness of our reduction layer identification approach. We provide additional numerical results in \cref{additionalnumericalresultsablation}.

\section{Conclusion}
In this work, we presented \textbf{PACT}, a method that addresses both visual token irrelevance and redundancy. \textbf{PACT} is a plug-and-play solution that does not require additional training. It does not rely on textual tokens for visual token reduction, making it well-suited for multi-turn conversations. Additionally, it operates independently of the visual encoder and connector architecture, making it broadly applicable across various Visual Language Models. Our results confirm that the number of visual tokens in Visual Language Models is unnecessarily large and provide valuable insights for effective token reduction. This opens the door for future work in designing more efficient connectors and architectures for VLMs.

\section{Acknowledgments}
This work received financial support from Crédit Agricole S.A. through the research chair with Ecole Polytechnique on  Trustworthy and Responsible AI. This work was granted access to the HPC resources of IDRIS under the allocation 2024-AD011014793R1 made by GENCI.

{
    \small
    \bibliographystyle{ieeenat_fullname}
    \bibliography{main}
}

\clearpage
\setcounter{page}{1}
\twocolumn[ 
\centering
\section*{ PACT: Pruning and Clustering-Based Token Reduction for Faster Visual Language Models    Supplementary Materials
}
\vspace{1em} 
] 

\appendix

\section{On the density peaks clustering algorithm}
\label{DPCAPPENDIC}
Density Peak Clustering (DPC) is a clustering algorithm that identifies cluster centers based on local density and the distance to points with higher density, denoted as \(\delta_i\). The density, \(\rho_i\), can be measured by counting the number of points within a cutoff distance \(d_c\) from \(\mathbf{u}_i\), or by using a Gaussian function where nearby points contribute more to the density, \(\rho_i = \sum_{j} \exp\left(-\left(\frac{d_{ij}}{d_c}\right)^2\right)\). Points with high \(\rho_i\) and \(\delta_i\) values are selected as cluster centers. This selection can be done by defining a threshold \(t\) and designating points as cluster centers where \(\rho_i \cdot \delta_i \geq t \times \max(\rho_i \cdot \delta_i)
\), or by selecting a fixed percentage. Other points are then assigned to the cluster of the nearest higher-density point, iterating from the highest to the lowest density. This process can create clusters of varying shapes, where the maximum distance between elements within a cluster can be extremely large. In extreme cases, the two farthest points in the input data can end up in the same cluster.
\section{DBDPC Characteristics}  
\label{Dbdpccharac}
This section aims to prove that \textbf{DBDPC} guarantees that: Each element’s distance to its assigned cluster center is at most \( d_c \) and that all cluster centers are at least \( d_c \) apart.

\noindent Assume, for contradiction, that at least one of the following statements is false:  
\begin{enumerate}
    \item There exists an element \( i \) assigned to a cluster such that its distance to the cluster center is greater than \( d_c \), i.e.,
    $d_{is} > d_c.$
    \item There exist two cluster centers \( s_1, s_2 \) such that their pairwise distance is at most \( d_c \), i.e., $d_{s_1s_2} \leq d_c.$
\end{enumerate}

\paragraph{Contradiction for Assumption 1}  
In \textbf{DBDPC}, each element \( i \) is assigned to its closest cluster center:
\[
s_i = \arg\min_{s \in C_{\text{centers}}} d_{is}.
\]
If \( d_{is} > d_c \) for a given center $s$, then we have \( d_{is'} > d_c \) for all centers. However, in the \textbf{DBDPC} selection process, an element is assigned as a cluster center if its minimum distance to already selected centers is over $d_c$. Thus, \( i \) should have been selected as a new cluster center, and its distance to the closest cluster center would be zero, which leads to a contradiction, proving that every element satisfies \( d_{is} \leq d_c \).  

\paragraph{Contradiction for Assumption 2}  
Assume, without loss of generality, that \( s_2 \) is chosen after \( s_1 \). By the center selection criterion, a new center \( s_2 \) is added only if:
\[
\min_{s \in C_{\text{centers}}} d_{s_2 s} > d_c.
\]
If \( d_{s_1s_2} \leq d_c \), then \( s_2 \) shouldn't be selected as a cluster center, which leads to a contradiction. Thus, no two centers can be closer than \( d_c \).

\noindent\textit{Inter-cluster distance upper-bound} :
Here we will refer to cosine similarity by $sim$. Let's $x$ and $y$ be two points in the same cluster, and $s$ their cluster center. Since each point $\mathbf{x}$ is within $d_c$ of its cluster center $\mathbf{s}$ and the distance used in the \textbf{DBDPC} algorithm is $1 -\operatorname{sim}$, we have $\operatorname{sim}(\mathbf{x}, \mathbf{s}) \ge 1 - d_c.$ We have from \cite{Trianglecosine}:
\[
\operatorname{sim}(\mathbf{x}, \mathbf{y}) 
\;\ge\; 
\operatorname{sim}(\mathbf{x}, \mathbf{s}) \cdot \operatorname{sim}(\mathbf{s}, \mathbf{y})
\;+\; 
m - 1,
\]
\[
\text{where } m = \min\!\Bigl\{\operatorname{sim}(\mathbf{x}, \mathbf{s})^2,\; \operatorname{sim}(\mathbf{s}, \mathbf{y})^2\Bigr\}.
\]
Using $\operatorname{sim}(\mathbf{x}, \mathbf{s}), \operatorname{sim}(\mathbf{s}, \mathbf{y}) \ge 1 - d_c$ we get
\[
\operatorname{sim}(\mathbf{x}, \mathbf{y}) 
\;\ge\; 
(1 - d_c)^2 
\;+\; 
(1 - d_c)^2 
\;-\; 
1
\;=\;
1 - 2\,d_c \,(2 - d_c).
\]
Finally, converting this back to the distance $d(\mathbf{x}, \mathbf{y}) = 1 - \operatorname{sim}(\mathbf{x}, \mathbf{y})$, we obtain:
\[
d(\mathbf{x}, \mathbf{y}) 
\;\le\; 
2 \, d_c \,(2 - d_c).
\]
Therefore, the intra-cluster distance in the \textbf{DBDPC} algorithm is bounded by $2\,d_c\,(2 - d_c)$.

\begin{algorithm}
\caption{Recursive Center Identification for DBDPC with Iterative Center Identification}
\label{algo:recursive_center_identification}
\begin{algorithmic}[0]
\INPUT Cutoff distance \( d_c \in \mathbb{R}^+ \), set of vectors \( \mathbf{U} = \{ \mathbf{u}_i \in \mathbb{R}^{d_l} \}_{i=1}^n \), density values \( \{ \rho_i \}_{i=1}^n \), distance matrix \( D = [d_{ij}] \), fallback threshold \( T > 0 \)
\OUTPUT Cluster center indices \( C_{\text{centers}} \)
\State Initialize cluster center set \( C_{\text{centers}} = \emptyset \)
\State Set the density of each point : \[
\rho_i = \text{argsort}(\{-\rho_j\}_{j=1}^n)[i]
\]

\While{\( \mathbf{U} \neq \emptyset \)}
    \State Compute \( \delta_i \) for all vectors \( \mathbf{u}_i \in \mathbf{U} \):
    \[
    \delta_i = \min_{\rho_j > \rho_i} d_{ij}
    \]
    \State Select cluster candidates:
    \[
    \mathbf{C}_{\text{new}} = \{ \mathbf{u}_i \in \mathbf{U} \mid \delta_i > d_c \}
    \]
    \State \( C_{\text{centers}} \gets C_{\text{centers}} \cup \mathbf{C}_{\text{new}} \)
    \State Update remaining vectors:
    \State \[
\mathbf{U} \gets \mathbf{U} \setminus \left( 
\mathbf{C}_{\text{new}} \cup 
\left\{ \mathbf{u}_k \in \mathbf{U} \mid 
\begin{array}{c}
\exists \mathbf{u}_i \in \mathbf{C}_{\text{new}} \\
\text{such that } d_{ik} \leq d_c 
\end{array} 
\right\} 
\right)
\]

    \If{\( |\mathbf{C}_{\text{new}}| < T \)}
        \State Order remaining vectors \( \mathbf{U} \) by decreasing \( \rho_i \):
        \State \( \mathbf{U} \gets \text{Sort}(\mathbf{U}, \text{key}=\rho_i, \text{order}=\text{descending}) \)
        \State Call Iterative Center Identification:
        \State \( C_{\text{centers}} \gets \text{IterativeCenterIdentification}(C_{\text{centers}}, \mathbf{U}, d_c) \)
        \State \Return \( C_{\text{centers}} \)
    \EndIf
\EndWhile
\State \Return \( C_{\text{centers}} \)

\vspace{1.0em}
\State \textbf{Function: Iterative Center Identification}
\State \textbf{Inputs:} Remaining vectors \( \mathbf{U} \) (ordered by \( \rho_i \)), current cluster center set \( C_{\text{centers}} \), cutoff distance \( d_c \)
\State \textbf{Outputs:} Updated cluster center indices \( C_{\text{centers}} \)

\ForAll{\( \mathbf{u}_i \in \mathbf{U} \)}
    \If{\( \min_{ \mathbf{u}_s \in C_{\text{centers}} } d_{is} > d_c \)}
        \State \( C_{\text{centers}} \gets C_{\text{centers}} \cup \{ \mathbf{u}_i \} \)
    \EndIf
\EndFor
\State \Return \( C_{\text{centers}} \)
\end{algorithmic}
\end{algorithm}

\section{A comparison between DBDPC and other clustering algorithms}
\label{Qualitative}

\noindent \textbf{Comparison between DBDPC and DPC}:  We note that, aside from using densities, \textbf{DBDPC} is fundamentally different from DPC. Please refer to \cref{DPCAPPENDIC} for a detailed explanation of the \textbf{DPC} algorithm. The center identification process in \textbf{DBDPC} results in two main characteristics with formal proof detailed in \cref{Dbdpccharac}. First, the distance between each element and its cluster center is below \( d_c \), which leads to inter-cluster distances being upper-bounded by \( 2d_c \times (2 - d_c) \). Additionally, the distance between cluster centers is lower-bounded by \( d_c \). These guarantees do not hold for DPC, leading to two drawbacks. Since inter-cluster distances are not controlled, merging these vectors may result in merging highly dissimilar vectors, leading to information loss. Also, in high-density regions, the distance between cluster centers becomes too small, making DPC ineffective in addressing information redundancy.  

\begin{figure}[!t]
    \centering
    \includegraphics[width=0.49\textwidth]{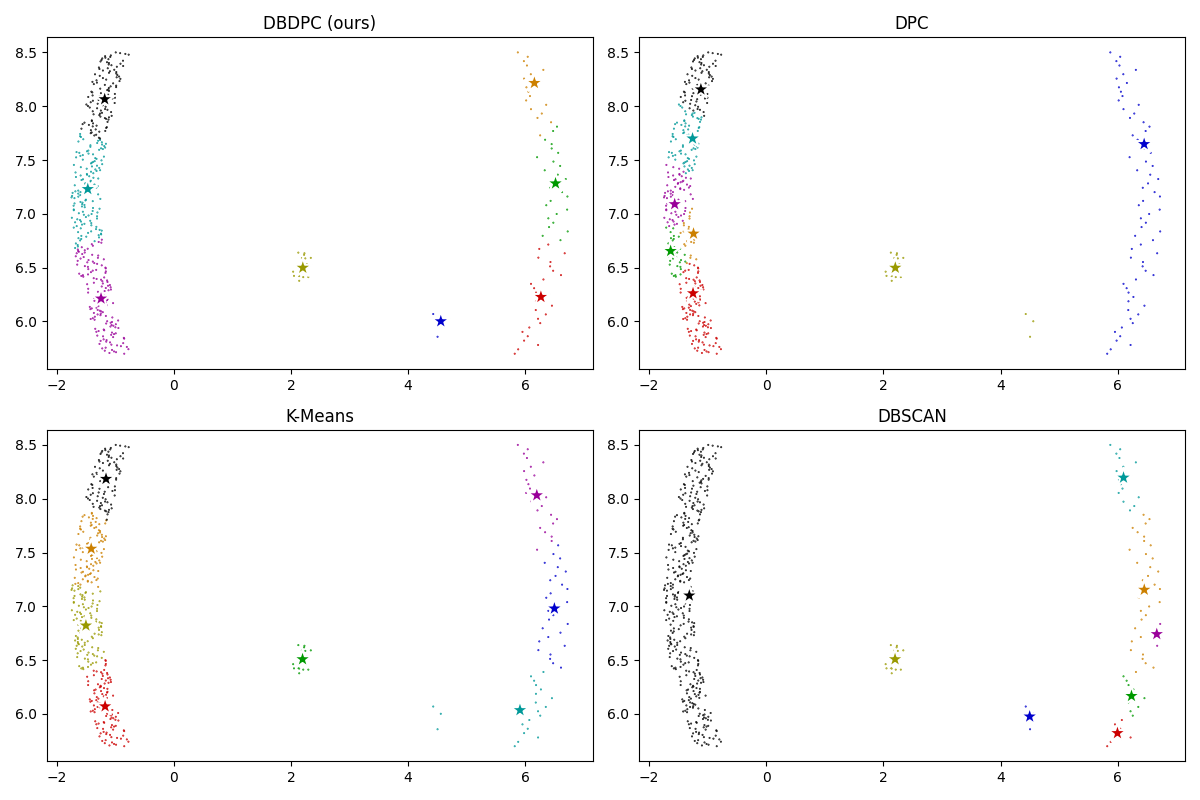}
    \caption{\textbf{An illustrative example of the difference in clustering characteristics between DBDPC and other clustering algorithms.} Two-dimensional points and the Euclidean distance were used for illustration purposes.}
    \label{clusteringillustration}
\vspace{-3pt} \end{figure}

\noindent\textbf{A Qualitative comparison}
 \Cref{clusteringillustration} presents the clustering results for \textbf{DBDPC}, DPC, DBSCAN, and K-Means on a predefined set of two-dimensional points. The figure shows that only \textbf{DBDPC} and DBSCAN identify isolated points as distinct clusters, a crucial feature for visual token reduction, as these points contain unique and thus potentially valuable information. We note that, for DBSCAN, these isolated points may be identified as noise, depending on the chosen hyperparameters. Moreover, \textbf{DBDPC} partitions both the left and right groups of points into the same number of clusters, maintaining consistency despite the higher density on the left side. In contrast, DPC tends to form a greater number of clusters in high-density regions while creating large clusters in low-density areas, whereas DBSCAN follows the opposite pattern, producing large clusters in high-density regions. In the context of visual token reduction, merging points within these large clusters can result in information loss, leading to performance degradation and making DPC and DBSCAN less suitable than \textbf{DBDPC} for this task. We note that the results presented in \cref{clusteringillustration} for DPC and DBSCAN may change when modifying the hyperparameters; however, the characteristics discussed above persist across different hyperparameter choices.

\section{Efficient center identification in DBDPC}
\label{effecientdbdpc}
\subsection{A recursive approach}

To enhance the efficiency of the \textbf{DBDPC} algorithm, we introduce a recursive center identification method that reduces computational overhead while maintaining clustering accuracy. In the \textbf{DBDPC} algorithm, vectors are processed in descending order of their local densities \(\rho_i\), and a vector \(\mathbf{u}_i\) is selected as a cluster center if it is farther than the cutoff distance \(d_c\) from all previously selected centers. Implementing this as described in  the algorithm requires sequentially iterating through all the vectors and checking distances to all previously selected centers, which does not fully leverage GPU parallelization capabilities. In the \textbf{DBDPC} algorithm, when two points have the same density, one is treated as if it has a higher density than the other, depending on the order of their processing. To replicate this behavior, we assign the density of each point to its rank as:
\[
\rho_i = \text{rank}_i = \text{argsort}(\{-\rho_j\}_{j=1}^n)[i]
\] Our accelerated method leverages the quantity \(\delta_i\), representing the minimum distance from vector \(\mathbf{u}_i\) to any higher-density vector:

\begin{equation}
\delta_i = \min_{\rho_j > \rho_i} d_{ij}
\label{delta_equation}
\end{equation}

If \(\delta_i > d_c\), then \(\mathbf{u}_i\) is selected as a cluster center because it is not within \(d_c\) of any higher-density vector, which are the only potential cluster centers that can be selected before \(d_{ij}\) in the \textbf{DBDPC} algorithm. In addition, any vector within \(d_c\) of a cluster center identified using \(\delta_i\) has a lower density than that center, as cluster centers identified using \(\delta_i\) are not within \(d_c\) of any higher-density vector. In the \textbf{DBDPC} algorithm, such a vector would not be chosen as a cluster center because it violates the distance condition relative to already selected centers. By identifying these vectors early, we can exclude them from further consideration as potential centers. We repeat this process recursively: after selecting cluster centers where \(\delta_i > d_c\) and excluding vectors within \(d_c\) of these centers, we process the remaining vectors. This recursion continues until the number of newly discovered cluster centers becomes small (e.g., less than 10). At that point, we fall back to the \textbf{DBDPC} method, processing the remaining vectors iteratively to ensure all potential centers are considered. This recursive approach reduces the number of iterations in the main loop and enhances parallelization, particularly on GPUs, by minimizing sequential computation. By leveraging $\delta_i$ and incorporating an early exclusion mechanism, the recursive center identification method reduces computational time while ensuring the same clustering results as the \textbf{DBDPC} algorithm. The recursive approach decreases the number of iterations and enhances GPU parallelization by minimizing sequential computation, making the algorithm more efficient for large datasets. The recursive center identification method is presented in \cref{algo:recursive_center_identification}. We note that in practice this recursive approach reduce the computational time of the \textbf{DBDPC} algorithm by around 3 times.

\subsection{Proof of correctness of the recursive approach} \label{correctness} To validate the correctness of the accelerated method, we demonstrate the following key points: selected centers are valid cluster centers, excluded vectors are not cluster centers and identifying remaining cluster centers is equivalent to identifying cluster centers on the reduced set.
\noindent Proving these points suffices to establish correctness, as the remaining vectors after the recursive steps are treated the same as in the \textbf{DBDPC} algorithm.

\noindent\textbf{Selected Centers Are Valid Cluster Centers} In the \textbf{DBDPC} algorithm, for any vector $\mathbf{u}_i$, only vectors with higher densities are considered for selection as cluster centers before $\mathbf{u}_i$. If $\mathbf{u}_i$ is not within $d_c$ of any higher-density vector (i.e., $\delta_i > d_c$) then the distance of $\mathbf{u}_i$ from any previously selected center cannot exceed the cutoff distance $d_c$. Consequently, $\mathbf{u}_i$ satisfies the condition for being a cluster center in the \textbf{DBDPC} algorithm, as it is farther than $d_c$ from all centers processed earlier.

\noindent\textbf{Excluded Vectors Are Not Cluster Centers} Vectors within \(d_c\) of a cluster center identified using \(\delta_i\) have lower densities than that center, as these centers are not within $d_c$ to any higher density point. In the \textbf{DBDPC} algorithm, such vectors would not be selected as cluster centers because they are within $d_c$ to an already selected center, violating the distance condition. Therefore, excluding these vectors early does not affect the selection of valid cluster centers.

\noindent\textbf{Identifying Remaining Cluster Centers is Equivalent to Identifying Cluster Centers on the Reduced Set} After selecting cluster centers where \(\delta_i > d_c\) and excluding vectors within \(d_c\) of these centers, we focus on the reduced set of remaining vectors for further processing. The critical observation is that the previously selected cluster centers are not within \(d_c\) of any vector in the reduced set. This is ensured by the exclusion step, where all vectors within \(d_c\) of these centers have been removed. Consequently, when identifying new cluster centers within the reduced set, we do not need to consider distances to the previously selected centers, as they cannot influence the selection due to their distance. Moreover, the vectors that have been excluded are not potential cluster centers themselves. Meaning that they can not influence the center selection process. This means that any vector satisfying $\delta >d_c$ in the reduced set, is actually not within $d_c$ to any higher density potential cluster center form the initial set, making it a cluster center.

\section{On the choice of Positional IDs for clustering algorithms}
\label{clusteringpositionalid}

In our work, we benchmark four clustering algorithms: agglomerative clustering \cite{agglomerative}, k-means \cite{kmeans}, Density Peaks Clustering (DPC) \cite{dpc}, and DBSCAN \cite{dbscan}.  For each algorithm, we use the key vectors for clustering, apply a cosine similarity-based distance (as in \textbf{DBDPC}), and evaluate two strategies: merging the hidden states within each cluster or selecting the cluster center as a representative token. We report the best-performing approach for each algorithm. Similar to \textbf{DBDPC}, we assign the position ID of the cluster center to the resulting vectors. However, apart from DPC, the other clustering algorithms do not explicitly provide a cluster center. For k-means and agglomerative clustering, we select the cluster center as the point closest to the average of all points in the cluster, using keys and cosine similarity. For DBSCAN, we experimented with choosing the point connected to the most other points within the cluster and found this approach to yield slightly better results, aligning better with the principles of DBSCAN. Thus, we adopted this strategy in our tests.
\vspace{-3pt}
\section{More about applying ToME to Visual Language Models}
\label{moreabouttome}
ToMe reduces the number of visual tokens at each layer of the transformer. For a given layer \(i\), the process starts by splitting the tokens into two distinct sets, A and B. Each token in set A is matched with its most similar counterpart in set B, using cosine similarity based on key vectors to determine the closest pairs. The top \(r_i\) pairs with the highest similarity are then selected for merging. Connected components from the matched pairs are combined into single vectors, where hidden states are averaged. It is important to note that each connected component contains exactly one element from set B, and when applying ToME to Visual Language Models, this element’s position ID is assigned to the merged token. In \cite{tome}, the number of visual tokens was reduced by a fixed quantity (\(r_i = r\)). However, this fixed reduction scheme cannot achieve more than a 50\% reduction unless no reduction is done at later layers when the number of tokens drops below \(r\), which goes against the gradual reduction strategy proposed in ToMe. To enable higher reduction ratios, we adopt a linearly decreasing scheduler, where the reduction is higher in early layers and decreases in later layers. This approach achieves a smaller average number of visual tokens across the network while still reducing the token count at each layer, allowing us to reach high reduction ratios effectively.

\section{Implementation details and hyper-parameters for PACT}
\label{implentationdetails}
\begin{figure}[]
    \centering
    \includegraphics[width=0.45\textwidth]{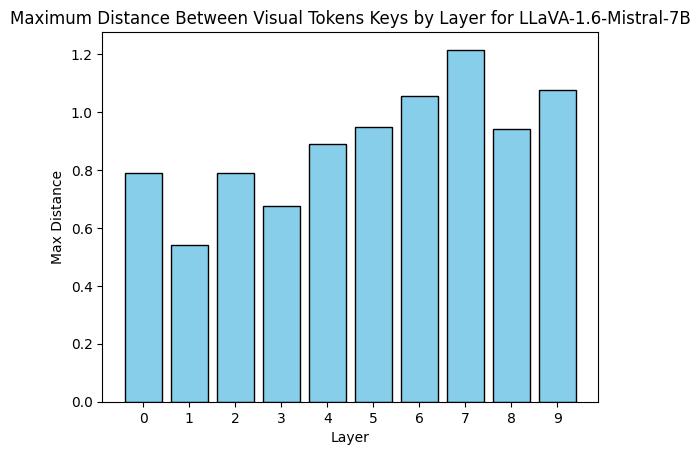}
\caption{\textbf{Illustration of the maximum distance between the keys of visual tokens for the first 10 layers of  LLaVA-1.6-Mistral-7B before the application of rotary embeddings.}}
    \label{plotdistancekeys1.6}
\vspace{-3pt} \end{figure}

\begin{figure}[]
    \centering
    \includegraphics[width=0.45\textwidth]{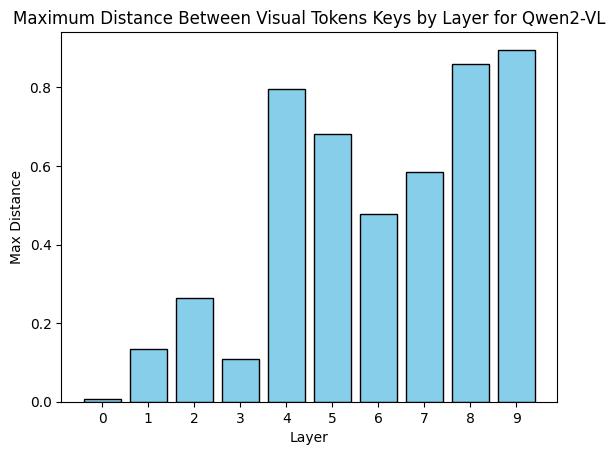}
\caption{\textbf{Illustration of the maximum distance between the keys of visual tokens for the first 10 layers of  Qwen2-VL-7B-Instruct before the application of rotary embeddings.}}
    \label{plotdistancekeysqwen}
\vspace{-3pt} \end{figure}

\begin{figure}[b!]
    \centering
    \includegraphics[width=0.45\textwidth]{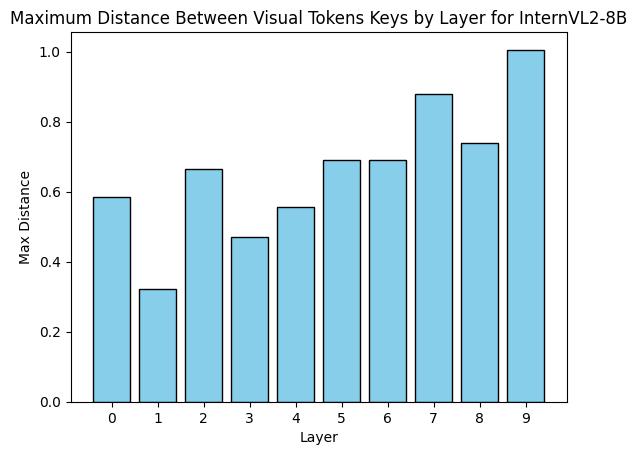}
\caption{\textbf{Illustration of the maximum distance between the keys of visual tokens for the first 10 layers of  InternVL2-8B before the application of rotary embeddings.}}
    \label{plotdistancekeysintern}
\vspace{-3pt} \end{figure}

For all experiments on LLaVA-OneVision-7B, we set \( d_n=2 \), \( \alpha=1.5 \), and \( L=4 \). While the optimal values of each parameter may vary depending on the dataset, we aim to evaluate the real-world effectiveness of our approach by using consistent values across all testing datasets. The results in \cref{tab:comparison_results} were obtained using \( d_c = 0.21 \) and \( \lambda = 0.55 \), while those in \cref{tab:performance_inference_summary} were obtained using \( d_c = 0.17 \) and \( \alpha = 0.7 \). Additionally, to demonstrate the performance of our approach at different reduction ratios, we vary \( d_c \) and \( \lambda \) and report the results. The values of the fixed parameters \( d_n \) and \( \alpha \) were chosen by performing a grid search on SeedBench \cite{SeedBench}, which is why we do not include SeedBench in the testing datasets. It is important to note that finding the optimal parameters for all testing datasets is not the focus of this study, as this would require extensive testing of different values for \( d_c \), \( \lambda \), \( L \), \( \alpha \), and \( d_n \) on all test sets. Such an approach would not accurately reflect the real-world performance of our method. Instead, we chose to only vary \( d_c \) and \( \lambda \) to evaluate the effectiveness of our approach at different reduction ratios. When testing on SeedBench, we found that a pruning ratio higher than 60\% harms performance. Therefore, we vary the pruning ratio between 10\% and 60\% and test across different values of \(d_c\). When testing \textbf{PACT} on LLaVA-1.6-Mistral-7B, Qwen2-VL-7B-Instruct and InternVL2-8B. We use the same values of $d_n$ and $\alpha$ as when testing on  LLaVA-OneVision-7B. We note that these hyperparameters may not be optimal; however, as we aim to test the generalizability of our approach, we opt to use the same hyperparameters across models. \Cref{plotdistancekeys1.6}, \Cref{plotdistancekeysqwen} and \Cref{plotdistancekeysintern} show the maximum distance between the keys at several layers of the language model for LLaVA-1.6-Mistral-7B, Qwen2-VL-7B-Instruct and InternVL2-8B. Following the same approach for  LLaVA-OneVision-7B, we choose $L=4$ for Qwen2-VL-7B-Instruct and $L=7$ for InternVL2-8B. We note that the choice of the reduction layer for InternVL2-8B is not as evident as for  LLaVA-OneVision-7B and Qwen2-VL-7B-Instruct, as the increase in maximum distance from one layer to the next is sometimes minimal, making it unclear which layer offers the best balance between accuracy and computational efficiency. However, since we do not aim to experimentally determine the optimal reduction layer, we end up choosing $L=7$, as the maximum distance between keys is increased by an acceptable amount between the seventh and eighth layer. Following the same approach we use $L=7$ for LLaVA-1.6-Mistral-7B.

\section{More about test datasets and used metrics}
\label{datasetsapendix}
\label{usedmetrics}

For evaluating the different approaches, we use LMMs-Eval \cite{LMMs-Eval} and aim to follow the same dataset splits and metrics as used in \cite{llavaone}. We detail the used splits and metrics in \cref{tab:dataset_splits}. Some datasets require evaluation using a GPT model through the OPENAI API or other closed-source models. However, for many datasets the version of the closed-source model used in evaluating LLaVA-OneVision in \cite{llavaone} is no longer available. So we use the latest version of GPT-4 for our assessments at the time of publication (gpt-4o-2024-08-06). We also observed that when calling a closed-source model like GPT-4 via an API, the responses are not fully deterministic, even with a temperature set to zero, introducing some noise into the evaluation metrics. To reduce this noise, we exclude all these datasets when testing across different reduction ratios. On the other hand, for \cref{tab:performance_inference_summary}, we exclude MMVet, Vibe-Eval, VideoChatGPT, MM-LiveBench, and LLaVA-Wilder as they have high inference times, which would dominate the throughput calculation.

\noindent For certain datasets, such as DocVQA, InfoVQA, and TextVQA, we use the validation split contrary to \cite{llavaone}. This choice allows us to test various reduction ratios and approaches without requiring submission to the test server, which would be impractical for extensive testing. For datasets requiring a test set submission (EgoSchema and PerceptionTest), where either the validation set is typically not used for evaluation or does not exist, we report the submission-based metrics evaluated directly on the test set. As explained above, for some datasets our evaluation setup differs from the one used for evaluating LLaVA-OneVision in \cite{llavaone}, which may result in variations in the reported results for this model on certain datasets. This is primarily due to the use of validation splits for DocVQA, InfoVQA, and TextVQA, as well as the reliance on GPT-based metrics for some datasets (a common practice for these benchmarks, making alternative evaluation difficult). Nevertheless, our comparisons remain fair, as the same evaluation procedure is consistently applied across all approaches and reduction ratios.

We note that when using reduction methods, results may include slight variations due to edge cases where distances or importance metrics for different vectors are equal. That’s why we report results based on the average of three different runs for each dataset.

Notably, when testing on Qwen2-VL-7B-Instruct without reduction, some datasets encountered GPU out-of-memory errors (MLVU, VideoMME, and ActivityNet Perception) which we excluded from the test set. Additionally, results on ScienceQA were quite low when tested without reduction (0.132), leading to its exclusion from testing as well. We note that, as we use LMM-Eval \cite{LMMs-Eval} for evaluation, results differ for some datasets from the officially reported results, as prompts are sometimes not formatted in the same manner. This observation also applies to InternVL2-8B.

\section{Additional numerical results}
\label{additionalnumericalresults}
\Cref{tab:comparison_clustering_1} and \cref{tab:comparison_clustering_2} show a comparison of \textbf{DBDPC} and various clustering algorithms for a reduction ratio of approximately 60\% on LLaVA-OneVision-7B across multiple datasets. The results demonstrate that \textbf{DBDPC} outperforms other clustering algorithms in visual token reduction for the majority of the datasets. Additionally, the tables show that the clustering process for \textbf{DBDPC} is significantly faster than that of other clustering algorithms.
\Cref{tab:comparison_pruning} presents a comparison of \textbf{EUTI}-based visual token pruning and FastV for a reduction ratio of approximately 60\% on LLaVA-OneVision-7B across various datasets. The results indicate that \textbf{EUTI} outperforms FastV on most datasets while also being more computationally efficient. 
\Cref{tab:ablation_keys_dbdpc} shows that using keys for distance calculations in DBDPC outperforms hidden states across the majority of the test datasets. Also, we present a comparison between \textbf{PACT} and other visual reduction techniques for InternVL2-8B, and LLaVA-1.6-Mistral-7B across different datasets in \cref{resultsintern}, and \cref{results_llava_1.6}.

\section{Ablation study : Additional numerical results}
\label{additionalnumericalresultsablation}
\Cref{tab:comparison_prune_merge} shows a comparison between \textbf{PACT}, \textbf{DBDPC}, and \textbf{EUTI} for a reduction ratio of approximately 70\%, applied on LLaVA-OneVision-7B. The results demonstrate that \textbf{PACT}, which combines both clustering and pruning, outperforms the other two methods that are either clustering-based or pruning-based across various datasets. More importantly, \textbf{DBDPC} and \textbf{EUTI} exhibit a significant drop in performance on some of the datasets, which is not the case for \textbf{PACT}. We note that numerical results for the ablation studies conducted on \textbf{DBDPC}, \textbf{EUTI}, and \textbf{PACT} can be found in \cref{tabablationDBDPC}, \cref{tabablationEUTI} and \cref{tabablationPACT}.


\begin{table*} \vspace{-3pt}
    \centering
    \caption{\small \textbf{Dataset Splits, Subsets, and Evaluation Metrics Used in Our Experiments.} Default indicates the use of the standard test split or cases where only one split/subset is available. The evaluation metrics employed are those commonly used for the respective datasets and generally the ones proposed in the official papers. For GPT-based scores (or any model-based scores), this means that a GPT model was used during evaluation, typically to extract answers from the generated output text, which are then matched with the ground truth to calculate accuracy using exact matches. When accuracy is reported, it generally implies that only an exact match is considered a correct answer.}
    \label{tab:dataset_splits}
    \begin{tabular}{@{} l l l l @{}} 
        \toprule
        \textbf{Dataset} & \textbf{Split} & \textbf{Subset} & \textbf{Evaluation Metric} \\
        \midrule
        VideoMME         & Default      & No subtitles   & Accuracy             \\
        MME              & Default      & Default        & MME Perception Score \\
        DocVQA           & Validation   & Default        & ANLS                 \\
        MLVU             & Default      & Default        & Accuracy             \\
        LLaVA-Interleave & Default      & Out-domain     & Accuracy             \\
        ChartQA          & Validation   & Default        & Relaxed Accuracy     \\
        MMBench          & Validation   & English        & GPT-based Score      \\
        MuirBench        & Default      & Default        & Accuracy             \\
        ScienceQA        & Default      & Vision only    & Accuracy             \\
        MMMU             & Validation   & Default        & Accuracy             \\
        AI2D             & Default      & Default        & Accuracy             \\
        InfographicVQA   & Validation   & Default        & ANLS                 \\
        MMStar           & Default      & Default        & Accuracy             \\
        ActivityNetQA    & Default      & Default        & GPT-based Score      \\
        MM-LiveBench     & Default      & 2406           & GPT-based Score      \\
        LLaVA-Wilder     & Default      & Small          & GPT-based Score      \\
        MathVerse        & Default      & Vision mini & GPT-based Score \\
        MathVista        & Default      & Testmini       & GPT-based Score      \\
        MMVet            & Default      & Default        & GPT-based Score      \\
        Vibe-Eval        & Default      & Default        & REKA-based Score     \\
        VideoChatGPT     & Default      & Default        & GPT-based Score      \\
        EgoSchema        & Default      & Default        & Submission           \\
        PerceptionTest   & Default      & Multiple Choice QA & Submission     \\
        TextVQA          & Validation   & Default        & Official metric  \\
        \bottomrule
    \end{tabular}
\vspace{-3pt} \end{table*}

\begin{table*}[!ht]
\centering
\caption{\small \textbf{Performance of \textbf{PACT} on LLaVA-OneVision-7B using $d_c=0.17$ and $\alpha=0.7$.} }
\label{tab:comparison_results_pact_only}
\resizebox{0.6\textwidth}{!}{
\begin{tabular}{@{} c c c c c @{}} 
    \toprule
    \textbf{Dataset} & \multicolumn{4}{c}{\textbf{PACT (Ours)}} \\
    \cmidrule(lr){2-5}
     & \textbf{Metric} & \textbf{Red. Ratio} & \textbf{Proc. Time} & \textbf{Algo. Time} \\
    \midrule
    \midrule
\textbf{VideoMME} & 57.7 & 69.2\% & 0.321 & 0.021 \\
\textbf{MME} & 1571.0 & 72.1\% & 0.226 & 0.017 \\
\textbf{DocVQA} & 85.4 & 71.1\% & 0.467 & 0.026 \\
\textbf{MLVU} & 64.8 & 69.2\% & 0.322 & 0.022 \\
\textbf{LLaVA-Interleave} & 62.2 & 72.2\% & 0.133 & 0.010 \\
\textbf{ChartQA} & 77.3 & 71.4\% & 0.309 & 0.019 \\
\textbf{MMBench} & 79.9 & 72.0\% & 0.134 & 0.010 \\
\textbf{MuirBench} & 42.4 & 70.9\% & 0.175 & 0.013 \\
\textbf{ScienceQA} & 93.5 & 72.0\% & 0.130 & 0.010 \\
\textbf{MMMU} & 48.8 & 72.6\% & 0.103 & 0.007 \\
\textbf{AI2D} & 81.2 & 72.5\% & 0.173 & 0.013 \\
\textbf{InfographicVQA} & 61.5 & 70.0\% & 0.403 & 0.023 \\
\textbf{MMStar} & 59.5 & 72.3\% & 0.147 & 0.011 \\
\textbf{ActivityNetQA} & 55.1 & 70.0\% & 0.409 & 0.029 \\
\textbf{MathVerse} & 17.1 & 76.0\% & 0.350 & 0.021 \\
\textbf{MathVista} & 62.1 & 73.0\% & 0.260 & 0.015 \\
\textbf{EgoSchema} & 60.0 & 69.1\% & 0.320 & 0.021 \\
\textbf{PerceptionTest} & 52.3 & 70.0\% & 0.301 & 0.023 \\
\textbf{TextVQA} & 75.5 & 69.2\% & 0.320 & 0.023 \\
    \bottomrule
\end{tabular}}
\end{table*}

\begin{table*} \vspace{-3pt}
\centering
\caption{\small \textbf{Comparison of \textbf{PACT} with FastV, VTW, and ToME applied on InternVL2-8B on Various Datasets.}}
\label{resultsintern}
\resizebox{\textwidth}{!}{
  \begin{tabular}{@{} c c c c c c c c c c c c c @{}} 
    \toprule
\textbf{Dataset} & \multicolumn{2}{c}{{\textbf{No Reduction}}} & \multicolumn{3}{c}{\textbf{PACT (Ours)}} & \multicolumn{2}{c}{\textbf{FastV}} & \multicolumn{2}{c}{\textbf{VTW}} & \multicolumn{2}{c}{\textbf{ToME}} \\
    \cmidrule(lr){2-3} \cmidrule(lr){4-6} \cmidrule(lr){7-8} \cmidrule(lr){9-10} \cmidrule(lr){11-12}
 & \textbf{Metric} & \textbf{Proc. Time} & \textbf{Metric} & \textbf{Red. Ratio} & \textbf{Proc. Time} & \textbf{Metric} & \textbf{Proc. Time} & \textbf{Metric} & \textbf{Proc. Time} & \textbf{Metric} & \textbf{Proc. Time} \\
    \midrule
\textbf{VideoMME} & 52.2 & 0.247 & \textbf{51.1} & 68.4\% & 0.151 & \textbf{51.1} & 0.155 & 51.0 & 0.142 & 50.2 & 0.190 \\
\textbf{MME} & 1621.0 & 0.171 & 1591.9 & 69.9\% & 0.121 & 1588.7 & 0.118 & \textbf{1627.0} & 0.111 & 1533.3 & 0.155 \\

\textbf{MLVU} & 50.6 & 0.439 & \textbf{49.7} & 68.8\% & 0.326 & 48.8 & 0.325 & 49.5 & 0.333 & 29.3 & 0.343 \\
\textbf{LLaVA-Interleave} & 40.0 & 0.390 & 39.0 & 71.2\% & 0.265 & \textbf{39.7} & 0.263 & 39.6 & 0.230 & 36.7 & 0.316 \\

\textbf{MMBench} & 81.9 & 0.161 & \textbf{80.4} & 70.4\% & 0.118 & 80.2 & 0.116 & 80.2 & 0.109 & 70.8 & 0.165 \\
\textbf{MuirBench} & 35.7 & 0.432 & 34.4 & 70.3\% & 0.249 & \textbf{35.6} & 0.258 & 33.7 & 0.210 & 32.7 & 0.296 \\
\textbf{ScienceQA} & 97.1 & 0.165 & \textbf{97.1} & 70.8\% & 0.118 & 95.8 & 0.116 & 95.7 & 0.109 & 89.9 & 0.151 \\
\textbf{MMMU} & 48.5 & 0.167 & \textbf{48.0} & 70.6\% & 0.126 & 47.7 & 0.126 & 47.8 & 0.119 & 47.5 & 0.156 \\
\textbf{AI2D} & 82.5 & 0.146 & \textbf{81.4} & 70.7\% & 0.112 & 78.5 & 0.110 & 79.6 & 0.105 & 74.4 & 0.142 \\

\textbf{MMStar} & 59.0 & 0.179 & \textbf{56.7} & 70.4\% & 0.186 & 54.2 & 0.184 & 53.4 & 0.352 & 55.1 & 0.156 \\

\textbf{PerceptionTest} & 57.7 & 0.300 & \textbf{56.8} & 66.0\% & 0.203 & 56.2 & 0.213 & 34.1 & 0.192 & 55.2 & 0.228 \\
\textbf{EgoSchema} & 54.0 & 0.240 & \textbf{53.7} & 67.0\% & 0.155 & 53.1 & 0.163 & 32.2 & 0.146 & 52.9 & 0.172 \\
\textbf{ActivityNet} & 51.7 & 0.240 & \textbf{51.3} & 66.0\% & 0.153 & 51.0 & 0.161 & 30.8 & 0.143 & 50.4 & 0.171 \\
\textbf{MM-LiveBench} & 68.0 & 3.075 & \textbf{67.3} & 68.0\% & 2.140 & 67.0 & 2.247 & 40.4 & 2.003 & 66.6 & 2.354 \\
    \bottomrule
  \end{tabular}}
\vspace{-3pt} \end{table*}

\begin{table*} \vspace{-3pt}
\centering
\caption{\small \textbf{Comparison of \textbf{PACT} with FastV, Prumerge, and Hired applied on LLaVA-1.6-Mistral-7B across multiple datasets.}}
\label{results_llava_1.6}
\resizebox{\textwidth}{!}{
  \begin{tabular}{@{} c c c c c c c c c c c c c @{}} 
    \toprule
\textbf{Dataset} & \multicolumn{2}{c}{{\textbf{No Reduction}}} & \multicolumn{3}{c}{\textbf{PACT (Ours)}} & \multicolumn{2}{c}{\textbf{FastV}} & \multicolumn{2}{c}{\textbf{Prumerge}} & \multicolumn{2}{c}{\textbf{Hired}} \\
    \cmidrule(lr){2-3} \cmidrule(lr){4-6} \cmidrule(lr){7-8} \cmidrule(lr){9-10} \cmidrule(lr){11-12}
 & \textbf{Metric} & \textbf{Proc. Time} & \textbf{Metric} & \textbf{Red. Ratio} & \textbf{Proc. Time} & \textbf{Metric} & \textbf{Proc. Time} & \textbf{Metric} & \textbf{Proc. Time} & \textbf{Metric} & \textbf{Proc. Time} \\
    \midrule
\textbf{MME} & 1500.0 & 0.237 & \textbf{1507.1} & 70.3\% & 0.159 & 1503.9 & 0.158 & 1485.4 & 0.166 & 1497.0 & 0.168 \\
\textbf{DocVQA} & 70.0 & 0.363 & \textbf{67.1} & 67.1\% & 0.284 & 64.5 & 0.281 & 48.8 & 0.293 & 65.8 & 0.295 \\
\textbf{ChartQA} & 52.9 & 0.332 & \textbf{49.3} & 70.1\% & 0.259 & 48.9 & 0.261 & 36.0 & 0.264 & 46.1 & 0.266 \\
\textbf{MMBench} & 68.2 & 0.226 & \textbf{68.0} & 71.9\% & 0.155 & 67.9 & 0.154 & 66.2 & 0.160 & 67.6 & 0.164 \\
\textbf{ScienceQA} & 73.0 & 0.197 & 72.7 & 71.5\% & 0.144 & \textbf{73.2} & 0.145 & 71.7 & 0.148 & 72.9 & 0.149 \\
\textbf{MMMU} & 34.2 & 0.239 & \textbf{34.9} & 71.5\% & 0.171 & 34.7 & 0.169 & 33.9 & 0.180 & 33.9 & 0.180 \\
\textbf{AI2D} & 67.5 & 0.233 & \textbf{67.5} & 70.9\% & 0.160 & 67.0 & 0.158 & 64.5 & 0.165 & 65.9 & 0.166 \\
\textbf{InfographicVQA} & 36.9 & 0.294 & \textbf{35.6} & 66.2\% & 0.226 & 33.4 & 0.229 & 31.9 & 0.236 & 31.6 & 0.236 \\
\textbf{MMStar} & 36.2 & 0.375 & \textbf{36.7} & 71.9\% & 0.350 & 36.6 & 0.400 & 35.1 & 0.345 & 35.9 & 0.345 \\
    \bottomrule
  \end{tabular}}
\vspace{-3pt} \end{table*}

\begin{table*} \vspace{-3pt}
\centering
\caption{\textbf{Comparison of DBDPC and Agglomerative Clustering Methods for a Reduction Ratio of approximately 60\% on LLaVA-OneVision-7B.}}
\label{tab:comparison_clustering_1}
\resizebox{\textwidth}{!}{ 
  \begin{tabular}{@{} c c c c c c c c c c c c c @{}} 
    \toprule
\textbf{Dataset} & \multicolumn{3}{c}{\textbf{DBDPC (ours)}} & \multicolumn{3}{c}{\textbf{Agg. (Single Linkage)}} & \multicolumn{3}{c}{\textbf{Agg. (Average Linkage)}} & \multicolumn{3}{c}{\textbf{Agg. (Complete Linkage)}} \\
    \cmidrule(lr){2-4} \cmidrule(lr){5-7} \cmidrule(lr){8-10} \cmidrule(lr){11-13}
    & \textbf{Metric} & \textbf{Proc. Time} & \textbf{Algo. Time} & \textbf{Metric} & \textbf{Proc. Time} & \textbf{Algo. Time} & \textbf{Metric} & \textbf{Proc. Time} & \textbf{Algo. Time} & \textbf{Metric} & \textbf{Proc. Time} & \textbf{Algo. Time} \\
    \midrule
\textbf{VideoMME} & 57.4 & 0.389 & 0.040 & 57.6 & 1.504 & 1.148 & 57.0 & 1.657 & 1.316 & 57.9 & 1.690 & 1.350 \\
\textbf{MME} & 1563.8 & 0.255 & 0.028 & 1554.1 & 0.994 & 0.738 & 1559.2 & 1.123 & 0.868 & 1563.0 & 1.151 & 0.897 \\
\textbf{DocVQA} & 84.7 & 0.530 & 0.044 & 83.6 & 1.899 & 1.379 & 84.4 & 2.185 & 1.662 & 84.3 & 2.308 & 1.777 \\
\textbf{MLVU} & 64.2 & 0.384 & 0.039 & 64.0 & 1.574 & 1.229 & 65.2 & 1.675 & 1.329 & 64.8 & 1.700 & 1.355 \\
\textbf{LLaVA-Interleave} & 62.1 & 0.151 & 0.016 & 62.0 & 0.425 & 0.277 & 61.5 & 0.446 & 0.298 & 61.4 & 0.446 & 0.298 \\
\textbf{ChartQA} & 76.0 & 0.366 & 0.031 & 74.5 & 1.151 & 0.798 & 75.8 & 1.253 & 0.910 & 75.8 & 1.277 & 0.930 \\
\textbf{MMBench} & 80.1 & 0.151 & 0.016 & 79.5 & 0.427 & 0.277 & 79.7 & 0.437 & 0.291 & 79.8 & 0.449 & 0.299 \\
\textbf{MuirBench} & 43.2 & 0.215 & 0.023 & 41.4 & 0.667 & 0.474 & 42.0 & 0.727 & 0.534 & 42.0 & 0.738 & 0.544 \\
\textbf{ScienceQA} & 94.7 & 0.147 & 0.015 & 94.8 & 0.394 & 0.250 & 94.7 & 0.416 & 0.271 & 94.7 & 0.413 & 0.269 \\
\textbf{MMMU} & 48.3 & 0.110 & 0.009 & 48.4 & 0.218 & 0.110 & 49.3 & 0.232 & 0.121 & 48.2 & 0.225 & 0.117 \\
\textbf{AI2D} & 80.7 & 0.202 & 0.022 & 80.8 & 0.667 & 0.472 & 80.6 & 0.748 & 0.551 & 80.1 & 0.753 & 0.557 \\
\textbf{InfographicVQA} & 61.6 & 0.528 & 0.046 & 57.1 & 1.608 & 1.181 & 59.8 & 1.818 & 1.394 & 59.8 & 1.870 & 1.436 \\
\textbf{MMStar} & 60.5 & 0.167 & 0.018 & 60.2 & 0.507 & 0.344 & 59.8 & 0.556 & 0.390 & 60.5 & 0.560 & 0.395 \\
    \bottomrule
  \end{tabular}}
\vspace{-3pt} \end{table*}

\begin{table*} \vspace{-3pt}
\centering
\caption{\textbf{Comparison of DBDPC, DBSCAN, DPC, and KMeans Clustering Methods for a Reduction Ratio of approximately 60\% on LLaVA-OneVision-7B.}}
\label{tab:comparison_clustering_2}
\resizebox{\textwidth}{!}{ 
  \begin{tabular}{@{} c c c c c c c c c c c c c c c @{}} 
    \toprule
\textbf{Dataset} & \multicolumn{3}{c}{\textbf{DBDPC (ours)}} & \multicolumn{3}{c}{\textbf{DBSCAN}} & \multicolumn{3}{c}{\textbf{DPC}} & \multicolumn{3}{c}{\textbf{KMeans}} \\
    \cmidrule(lr){2-4} \cmidrule(lr){5-7} \cmidrule(lr){8-10} \cmidrule(lr){11-13}
    & \textbf{Metric} & \textbf{Proc. Time} & \textbf{Algo. Time} & \textbf{Metric} & \textbf{Proc. Time} & \textbf{Algo. Time} & \textbf{Metric} & \textbf{Proc. Time} & \textbf{Algo. Time} & \textbf{Metric} & \textbf{Proc. Time} & \textbf{Algo. Time} \\
    \midrule
\textbf{VideoMME} & 57.4 & 0.389 & 0.040 & 57.4 & 0.394 & 0.046 & 56.9 & 0.729 & 0.392 & 57.3 & 1.725 & 1.383 \\
\textbf{MME} & 1563.8 & 0.255 & 0.028 & 1560.3 & 0.274 & 0.036 & 1549.9 & 0.637 & 0.380 & 1549.9 & 1.254 & 0.999 \\
\textbf{DocVQA} & 84.7 & 0.530 & 0.044 & 84.2 & 0.533 & 0.044 & 83.0 & 0.950 & 0.442 & 79.6 & 2.059 & 1.544 \\
\textbf{MLVU} & 64.2 & 0.384 & 0.039 & 64.2 & 0.391 & 0.048 & 64.2 & 0.727 & 0.382 & 64.6 & 1.725 & 1.377 \\
\textbf{LLaVA-Interleave} & 62.1 & 0.151 & 0.016 & 60.4 & 0.159 & 0.026 & 63.9 & 0.258 & 0.121 & 62.3 & 0.711 & 0.566 \\
\textbf{ChartQA} & 76.0 & 0.366 & 0.031 & 75.2 & 0.369 & 0.034 & 75.2 & 0.758 & 0.415 & 74.2 & 1.399 & 1.059 \\
\textbf{MMBench} & 80.1 & 0.151 & 0.016 & 78.1 & 0.153 & 0.020 & 79.5 & 0.326 & 0.179 & 79.9 & 0.702 & 0.552 \\
\textbf{MuirBench} & 43.2 & 0.215 & 0.023 & 42.4 & 0.219 & 0.028 & 42.0 & 0.466 & 0.273 & 42.9 & 0.955 & 0.763 \\
\textbf{ScienceQA} & 94.7 & 0.147 & 0.015 & 91.2 & 0.150 & 0.024 & 94.3 & 0.251 & 0.117 & 93.4 & 0.661 & 0.518 \\
\textbf{MMMU} & 48.3 & 0.110 & 0.009 & 47.8 & 0.130 & 0.030 & 48.3 & 0.187 & 0.078 & 48.2& 0.500 & 0.391 \\
\textbf{AI2D} & 80.7 & 0.202 & 0.022 & 79.2 & 0.202 & 0.022 & 80.3 & 0.455 & 0.264 & 81.1 & 1.062 & 0.860 \\
\textbf{InfographicVQA} & 61.6 & 0.528 & 0.046 & 54.0 & 0.531 & 0.052 & 56.6 & 0.975 & 0.547 & 57.8 & 1.780 & 1.357 \\
\textbf{MMStar} & 60.5 & 0.167 & 0.018 & 56.6 & 0.179 & 0.028 & 60.6 & 0.376 & 0.213 & 60.2 & 0.828 & 0.661 \\
    \bottomrule
  \end{tabular}}
\vspace{-3pt} \end{table*}
\begin{table*} \vspace{-3pt}
\centering
\caption{\textbf{Comparison of EUTI-based visual tokens pruning and FastV for a Reduction Ratio of approximately 60\% on LLaVA-OneVision-7B.}}
\label{tab:comparison_pruning}{ 
  \begin{tabular}{@{} c c c c c c c @{}} 
    \toprule
\textbf{Dataset} & \multicolumn{3}{c}{\textbf{EUTI (Ours)}} & \multicolumn{3}{c}{\textbf{FastV}} \\
    \cmidrule(lr){2-4} \cmidrule(lr){5-7}
    & \textbf{Metric} & \textbf{Proc. Time} & \textbf{Algo. Time} & \textbf{Metric} & \textbf{Proc. Time} & \textbf{Algo. Time} \\
    \midrule
\textbf{VideoMME} & 58.4 & 0.351 & 0.005 & 57.6 & 0.381 & 0.040 \\
\textbf{MME} & 1560.0 & 0.256 & 0.004 & 1570.7 & 0.283 & 0.025 \\
\textbf{DocVQA} & 86.5 & 0.521 & 0.005 & 85.3 & 0.559 & 0.032 \\
\textbf{MLVU} & 64.3 & 0.355 & 0.004 & 63.1 & 0.391 & 0.040 \\
\textbf{LLaVA-Interleave} & 58.9 & 0.140 & 0.003 & 59.7 & 0.152 & 0.007 \\
\textbf{ChartQA} & 78.6 & 0.344 & 0.004 & 78.0 & 0.363 & 0.016 \\
\textbf{MMBench} & 80.2 & 0.142 & 0.003 & 79.2 & 0.151 & 0.005 \\
\textbf{MuirBench} & 40.0 & 0.191 & 0.003 & 40.8 & 0.204 & 0.009 \\
\textbf{ScienceQA} & 93.6 & 0.137 & 0.003 & 92.3 & 0.149 & 0.006 \\
\textbf{MMMU} & 48.8 & 0.101 & 0.002 & 47.3 & 0.110 & 0.003 \\
\textbf{AI2D} & 81.1 & 0.191 & 0.003 & 80.3 & 0.202 & 0.009 \\
\textbf{InfographicVQA} & 63.0 & 0.425 & 0.005 & 60.3 & 0.473 & 0.040 \\
\textbf{MMStar} & 59.6 & 0.159 & 0.003 & 59.6 & 0.170 & 0.007 \\
    \bottomrule
  \end{tabular}}
\vspace{-3pt} \end{table*}

\begin{table*} \vspace{-3pt}
\centering
\caption{\textbf{Comparison of PACT with Standalone Methods: EUTI-based Visual Token Pruning and DBDPC Clustering Algorithm for a Reduction Ratio of approximately 70\%, applied on LLaVA-OneVision-7B.}}
\label{tab:comparison_prune_merge}
\resizebox{\textwidth}{!}{
  \begin{tabular}{@{} c c c c c c c c c c @{}} 
    \toprule
    \textbf{Dataset} & \multicolumn{3}{c}{\textbf{PACT}} & \multicolumn{3}{c}{\textbf{DBDPC}} & \multicolumn{3}{c}{\textbf{EUTI}} \\
    \cmidrule(lr){2-4} \cmidrule(lr){5-7} \cmidrule(lr){8-10}
    & \textbf{Metric} & \textbf{Proc. Time} & \textbf{Algo. Time} & \textbf{Metric} & \textbf{Proc. Time} & \textbf{Algo. Time} & \textbf{Metric} & \textbf{Proc. Time} & \textbf{Algo. Time} \\
    \midrule
    \textbf{VideoMME} & 57.5 & 0.321 & 0.021 & 57.3 & 0.342 & 0.040 & 58.4 & 0.305 & 0.005 \\
    \textbf{MME} & 1558.7 & 0.226 & 0.017 & 1543.7 & 0.243 & 0.028 & 1595.9 & 0.213 & 0.004 \\
    \textbf{DocVQA} & 84.3 & 0.467 & 0.026 & 82.5 & 0.500 & 0.044 & 85.3 & 0.456 & 0.005 \\
    \textbf{MLVU} & 64.6 & 0.322 & 0.022 & 63.9 & 0.358 & 0.039 & 64.4 & 0.291 & 0.004 \\
    \textbf{LLaVA-Interleave} & 63.9 & 0.133 & 0.010 & 62.6 & 0.149 & 0.016 & 57.1 & 0.127 & 0.003 \\
    \textbf{ChartQA} & 77.2 & 0.311 & 0.019 & 75.1 & 0.333 & 0.031 & 78.2 & 0.292 & 0.004 \\
    \textbf{MMBench} & 80.2 & 0.134 & 0.010 & 79.7 & 0.147 & 0.016 & 79.6 & 0.128 & 0.003 \\
    \textbf{MuirBench} & 42.8 & 0.175 & 0.013 & 43.2 & 0.211 & 0.023 & 39.9 & 0.164 & 0.003 \\
    \textbf{ScienceQA} & 93.6 & 0.130 & 0.010 & 93.8 & 0.142 & 0.015 & 92.2 & 0.123 & 0.003 \\
    \textbf{MMMU} & 48.9 & 0.103 & 0.007 & 47.2 & 0.109 & 0.009 & 48.9 & 0.096 & 0.002 \\
    \textbf{AI2D} & 80.6 & 0.173 & 0.013 & 80.5 & 0.191 & 0.022 & 79.9 & 0.164 & 0.003 \\
    \textbf{InfographicVQA} & 61.9 & 0.403 & 0.023 & 58.8 & 0.465 & 0.046 & 60.4 & 0.360 & 0.005 \\
    \textbf{MMStar} & 59.5 & 0.147 & 0.011 & 59.5 & 0.163 & 0.018 & 59.2 & 0.140 & 0.003 \\
    \bottomrule
  \end{tabular}}
\vspace{-3pt} \end{table*}

\begin{table*} \vspace{-3pt}
\centering
\caption{\textbf{Ablation Studies on DBDPC-based visual token reduction for a Reduction Ratio of approximately 60\% on LLaVA-OneVision-7B.} We report only the metrics, as processing time is similar across different approaches. When ablating the Center Position-IDs assignment, we reorder the hidden states based on the mean of the Position-IDs of the elements in each cluster and then assign position IDs sequentially.}
\label{tabablationDBDPC}{
  \begin{tabular}{@{} c c c c c @{}} 
    \toprule
      & \textbf{DBDPC} & \textbf{w/o Center Position-IDs assignment} & \textbf{w/o Proportional Attention} & \textbf{w/o Merging} \\
    \midrule
    \textbf{VideoMME} & 57.4 & 58.0 & 57.9 & 57.5 \\
    \textbf{MME} & 1563.8 & 1539.3 & 1523.8 & 1476.9 \\
    \textbf{DocVQA} & 84.7 & 28.2 & 84.2 & 83.1 \\
    \textbf{MLVU} & 64.2 & 61.2 & 63.9 & 63.5 \\
    \textbf{LLaVA-Interleave} & 62.1 & 69.6 & 63.2 & 63.6 \\
    \textbf{ChartQA} & 76.0 & 24.8 & 76.0 & 74.4 \\
    \textbf{MMBench} & 80.1 & 76.1 & 80.1 & 79.6 \\
    \textbf{MuirBench} & 43.2 & 26.5 & 43.2 & 44.0 \\
    \textbf{ScienceQA} & 94.7 & 67.4 & 94.2 & 93.6 \\
    \textbf{MMMU} & 48.3 & 34.5 & 47.6 & 48.2 \\
    \textbf{AI2D} & 80.7 & 43.0 & 80.4 & 79.9 \\
    \textbf{InfographicVQA} & 61.6 & 17.8 & 59.8 & 58.7 \\
    \textbf{MMStar} & 60.5 & 58.9 & 59.6 & 59.1 \\
    \bottomrule
  \end{tabular}
}
\vspace{-3pt} \end{table*}

\begin{table*} \vspace{-3pt}
\centering
\caption{\textbf{Ablation Studies on the EUTI-based Visual Token Pruning for a Reduction Ratio of approximately 70\%, applied on LLaVA-OneVision-7B.} We report only the metrics, as processing time is similar across different approaches.}
\label{tabablationEUTI}{
  \begin{tabular}{@{} c c c c @{}} 
    \toprule
    \textbf{Dataset} & \textbf{EUTI} & \textbf{EUTI w/o Norm} & \textbf{Norm (EUTI w/o Global Query)} \\
    \midrule
    \textbf{VideoMME} & 58.4 & 57.6 & 56.6 \\
    \textbf{MME} & 1595.9 & 1573.4 & 1576.5 \\
    \textbf{DocVQA} & 85.3 & 85.1 & 79.7 \\
    \textbf{MLVU} & 64.3 & 63.0 & 63.1 \\
    \textbf{LLaVA-Interleave} & 57.1 & 57.9 & 52.9 \\
    \textbf{ChartQA} & 78.2 & 76.4 & 76.7 \\
    \textbf{MMBench} & 79.6 & 79.4 & 79.4 \\
    \textbf{MuirBench} & 40.0 & 40.5 & 39.6 \\
    \textbf{ScienceQA} & 92.2 & 91.8 & 93.5 \\
    \textbf{MMMU} & 48.9 & 49.3 & 49.2 \\
    \textbf{AI2D} & 79.9 & 79.9 & 79.7 \\
    \textbf{InfographicVQA} & 60.4 & 60.1 & 49.3 \\
    \textbf{MMStar} & 59.2 & 57.4 & 59.2 \\
    \bottomrule
  \end{tabular}
}
\vspace{-3pt} \end{table*}

\begin{table*} \vspace{-3pt}
\centering
\caption{\textbf{Ablation Study on Pruned Tokens Recovery for a Reduction Ratio of approximately 70\%.} We remove the token recovery step, which is equivalent to Setting $\alpha$ to Zero. We report only the metrics, as processing time is similar across both approaches.}
\label{tabablationPACT}{
  \begin{tabular}{@{} c c c @{}} 
    \toprule
    \textbf{Dataset} & \textbf{PACT} & \textbf{PACT w/o Pruned-Token Recovery} \\
    \midrule
    \textbf{VideoMME} & 57.6 & 57.4 \\
    \textbf{MME} & 1556.7 & 1576.3 \\
    \textbf{DocVQA} & 84.3 & 84.3 \\
    \textbf{MLVU} & 64.6 & 64.2 \\
    \textbf{LLaVA-Interleave} & 63.9 & 59.6 \\
    \textbf{ChartQA} & 76.4 & 76.4 \\
    \textbf{MMBench} & 79.9 & 79.8 \\
    \textbf{MuirBench} & 42.8 & 42.2 \\
    \textbf{ScienceQA} & 93.3 & 93.6 \\
    \textbf{MMMU} & 48.5 & 48.5 \\
    \textbf{AI2D} & 80.6 & 80.6 \\
    \textbf{InfographicVQA} & 61.9 & 61.3 \\
    \textbf{MMStar} & 75.1 & 74.9 \\
    \bottomrule
  \end{tabular}
}
\vspace{-3pt} \end{table*}

\begin{table*} \vspace{-3pt}
\centering
\caption{\textbf{Ablation Study on Keys Utilization in DBDPC for a Reduction Ratio of approximately 60\%.} Metrics are reported, as processing time is similar across both configurations.}
\label{tab:ablation_keys_dbdpc}{
  \begin{tabular}{@{} c c c @{}} 
    \toprule
    \textbf{Dataset} & \textbf{DBDPC} & \textbf{DBDPC w/o Keys} \\
    \midrule
    \textbf{VideoMME} & 57.40 & 57.22 \\
    \textbf{MME} & 1563.80 & 1526.18 \\
    \textbf{DocVQA} & 84.70 & 80.50 \\
    \textbf{MLVU} & 64.20 & 64.60 \\
    \textbf{LLaVA-Interleave} & 62.10 & 60.80 \\
    \textbf{ChartQA} & 76.00 & 68.80 \\
    \textbf{MMBench} & 80.10 & 79.21 \\
    \textbf{MuirBench} & 43.20 & 41.40 \\
    \textbf{ScienceQA} & 94.70 & 91.90 \\
    \textbf{MMMU} & 48.30 & 47.90 \\
    \textbf{AI2D} & 80.70 & 79.10 \\
    \textbf{InfographicVQA} & 61.6 & 56.70 \\
    \textbf{MMStar} & 60.50 & 58.40 \\
    \bottomrule
  \end{tabular}
}
\vspace{-3pt} \end{table*}

\end{document}